\documentclass[10pt,journal,compsoc]{IEEEtran}

\ifCLASSOPTIONcompsoc
  \usepackage[nocompress]{cite}
	\usepackage{mathtools}
 \usepackage{amsthm}
  \newtheoremstyle{dotless}{}{}{\itshape}{}{\bfseries}{}{ }{}
  \theoremstyle{dotless}
  \newtheorem*{thm}{Theorem}
	
	\newtheorem*{cor}{Corollary}
	
\usepackage{cleveref}
\usepackage{calc}
\usepackage{algpseudocode}
\usepackage{algorithm}
\usepackage{mathtools}
\usepackage{amsmath}
\usepackage{cleveref}
\usepackage{calc}
\usepackage{amssymb}
\usepackage{easybmat}

\else
  \usepackage{cite}
\fi
\ifCLASSINFOpdf
\else
\fi

\usepackage{algorithm}
\usepackage{algpseudocode}
\begin{document}
%
\title{Robust One-Class Kernel Spectral Regression}
%
%
%
%

\author{Shervin~Rahimzadeh~Arashloo~\IEEEmembership{}and~Josef~Kittler,~\IEEEmembership{Life~member,~IEEE}
\IEEEcompsocitemizethanks{\IEEEcompsocthanksitem S.R. Arashloo is with the department
of computer engineering, faculty of engineering, Bilkent university, Ankara, Turkey, 06800.\protect\\
E-mail: S.rahimzadeh@cs.bilkent.edu.tr
\IEEEcompsocthanksitem J. Kittler is with CVSSP, department of electronic engineering, faculty of engineering and physical sciences, university of Surrey, Guildford, Surrey, UK, GU2 7XH.\protect\\ E-mail: J.kittler@surrey.ac.uk}
\thanks{}}

%
%

\markboth{}%
{}
%



\IEEEtitleabstractindextext{%
\begin{abstract}
The kernel null-space technique and its regression-based formulation (called one-class kernel spectral regression, a.k.a. OC-KSR) is known to be an effective and computationally attractive one-class classification framework. Despite its outstanding performance, the applicability of kernel null-space method is limited due to its susceptibility to possible training data corruptions and inability to rank training observations according to their conformity with the model. This work addresses these shortcomings by studying the effect of regularising the solution of the null-space kernel Fisher methodology in the context of its regression-based formulation (OC-KSR). In this respect, first, the effect of a Tikhonov regularisation in the Hilbert space is analysed where the one-class learning problem in presence of contaminations in the training set is posed as a sensitivity analysis problem. Next, driven by the success of the sparse representation methodology, the effect of a sparsity regularisation on the solution is studied. For both alternative regularisation schemes, iterative algorithms are proposed which recursively update label confidences and rank training observations based on their fit with the model. Through extensive experiments conducted on different data sets, the proposed methodology is found to enhance robustness against contamination in the training set as compared with the baseline kernel null-space technique as well as other existing approaches in a one-class classification paradigm while providing the functionality to rank training samples effectively.
\end{abstract}

}

\maketitle

\IEEEdisplaynontitleabstractindextext

%
\IEEEpeerreviewmaketitle

\ifCLASSOPTIONcompsoc
\IEEEraisesectionheading{\section{Introduction}\label{sec:introduction}}
\else
\section{Introduction}
\label{sec:introduction}

\fi

%
%
%
%
\IEEEPARstart{W}{hile} a wide variety of pattern classification problems enjoy an abundance of data for system design, there exist other applications which suffer from imbalanced or unrepresentative training set. This is manifested in applications where the cost associated with collecting samples corresponding to a specific class is prohibitively high, or samples take unpredictable novel forms at the test time. There also exist other scenarios where the recognition task is inherently open, leading to inaccurately defined class(es). In these situations, a particular class may not be very well represented by the available training samples or be totally void of samples, causing a degradation in the performance of the conventional multi/binary-class classifiers. An alternative and more effective approach in such cases is offered by the one-class classification (OCC) paradigm \cite{Chandola:2009:ADS:1541880.1541882}. One-class classification aims to identify patterns which conform to a specific behaviour/condition, recognised as the target class, and distinguish them from all other non-target objects. It differs from the the conventional multi/binary-class formulation in that it primarily uses samples from a single class for training. More concretely, assume $X = \{x_1, \dots, x_{n} \}$ to be a set of observations where $x_i\in\mathbb{R}^d$ is a realisation of a multivariate random variable $x$ characterised by the target probability density function $p(x)$. One-class learning tries to specify the support domain of $p(x)$ using a one-class classifier $h(z)$ as
\begin{equation}
 h(z) = \lceil f(z)\geq\tau\rceil=\left\{
  \begin{array}{ll}
    1 \textrm{\hspace{.5cm}} z\in \mathcal{T}\\
    0 \textrm{\hspace{.5cm} otherwise}
  \end{array} \right.
\end{equation}
\noindent where $\mathcal{T}$ denotes the target class and function $f(.)$ encodes the similarity of an observation $z$ to $\mathcal{T}$, while $\lceil . \rceil$ stands for the Iverson brackets. Parameter $\tau$ controls the fraction of observations which lie within the support domain of the target distribution. One-class classification is typically believed to be more difficult than its multi/binary-class counterpart due to a lack of non-target training samples, impeding the estimation of a decision boundary between the target and non-target observations. Yet, it forms the basis of a variety of different applications including intrusion detection \cite{6846360}, novelty detection \cite{BEGHI20141953}, fault detection in safety-critical systems \cite{4694106}, fraud detection \cite{Kamaruddin:2016:CCF:2980258.2980319}, insurance \cite{7435726}, health care \cite{6566012}, surveillance \cite{4668357}, etc. In order to identify a pattern as normal or novel, a one-class classifier is trained on a training set consisting of (typically) normal samples and then used to gauge the similarity of a test sample to those previously observed in the training set. In this case, the \textit{generalisation capability} of the one-class learner on new data plays a central role. In a real-life one-class learning problem, however, not all observations in the training set would conform to the model equally. In practice, the training set may be corrupted and incorporate noisy or non-target observations which may degrade performance, emphasising the requirement for mechanisms to spot such samples. Identification of the degree of normality of samples in  a given set is also useful in its own right for specific applications such as ranking and retrieving items for a given query in a database, pruning contaminated data sets as well as in decision threshold setting. Other possible uses may include providing the functionality to enhance system performance by refining the decision boundary using the identified counter-examples. Consequently, a \textit{ranking} of the set of training observations according to their fit with the model is highly desirable in a one-class learning paradigm. 

While there exist a variety of different OCC methods and fine-grained categorisation of such methods exists \cite{khan_madden_2014,30e27ea86,PIMENTEL2014215}, they can be roughly identified as either generative or non-generative \cite{6636290}. Generative approaches include a model for data generation whereas non-generative methods lack a transparent link to the data. Among well-known instances of the generative approaches are the parametric and non-parametric density estimation methods \cite{Tax:2001:COC:648055.744087,10.1137/1.9781611973440.67,HOFFMANN2007863}, neural-network based approaches \cite{Japkowicz:1999:CLA:929980,7492368}, sparse representation methods \cite{7984788,7393467}, etc. Notable representatives of the non-generative methods include support vector-based approaches \cite{Tax2004,Scholkopf:2001:ESH:1119748.1119749}, convex hull methods \cite{10.1007/978-3-642-21557-5_13,8125573}, cluster approaches \cite{10.1007/978-1-4471-1599-1_110} and subspace techniques  \cite{NIPS2004_2656,Roth:2006:KFD:1117520.1117529,6619277,6857384,6384805}. As a result of the particular importance given to classification, rather than generative process modelling, the non-generative methods are typically believed to yield better classification performance. Among other non-generative methods, the kernel null-space approach \cite{DBLP:journals/corr/abs-1807-01085,6619277,8099922} is known to be a very successful methodology for one-class classification. In particular, it has been found to perform better than many state-of-the-art one-class classification techniques by operating on a non-linear projection function corresponding to the optimal Fisher criterion for classification. More specifically, in this methodology, a discriminative feature subspace is inferred such that observations from a specific class are mapped onto a \textit{single} point, as a result of which a zero within-class scatter is obtained. On the other hand, a positive between-class scatter is obtained since samples corresponding to different classes are mapped onto distinct points in the feature subspace. Although the superiority of this approach over some other alternatives has been confirmed in different studies \cite{DBLP:journals/corr/abs-1807-01085,6619277,8099922}, nevertheless, in practice, it suffers from a number of limitations. In a one-class learning paradigm it is natural to expect that not all training samples equally agree with the model inferred. In extreme cases, a fraction of training observations might correspond to corrupted observations or pure counter-examples. As such, it is desirable to rank training observations according to their conformity with the inferred model. A ranking of training observations may then allow one to enhance performance by discarding noisy observations from the training set or using incompatible samples as counter-examples to refine decision boundaries and thus enhance robustness against contaminations in the training set. In addition, a ranking of training data also facilitates setting a decision threshold for the system to achieve (approximately) a desired error rate on the test set. Furthermore, there exist applications where one is only interested in ranking observations in a given data set in a completely unsupervised fashion. However, as in the kernel null-space methodology all training samples corresponding to a particular class are projected onto the same point in an optimal feature subspace, there exists no straightforward mechanism to gauge the compatibility of (and subsequently rank) training samples with the underlying model. As will be demonstrated in the experimental evaluation section, the kernel null-space technique is sensitive to noisy observations in the training set which can seriously degrade its performance in practical settings.

\subsection{Overview of the proposed approach}
This work addresses the aforementioned limitations of the kernel null-space methodology by: 1-allowing observation label to vary and thus providing soft confidences in contrast to fixed hard labels in the original kernel null-space technique, facilitating a ranking of training observations; 2-studying the effect of different regularisation techniques in a regression framework to deal with contaminations in the training set of a one-class classifier and improve its robustness; and 3-refining the decision boundary of a one-class classifier by automatically detecting contaminations in the training set and utilising them as counter-examples. In terms of the first contribution, an alternating minimisation approach is proposed where the solution in the Hilbert space and label confidences are optimised concurrently. The proposed approach thus provides a soft label assignment to training samples which facilitates observation ranking. Next, two alternative regularisation techniques based on Tikhonov and sparsity are studied and shown they may improve system robustness against contaminations in the training set in the context of one-class kernel null-space formulation. And, finally, in terms of the third contribution, a variant of the proposed approach is presented to make use of possible information regarding the number of non-target samples in the training set. Such information is naturally incorporated into the learning mechanism and found to enhance system performance by detecting outliers in the training set and utilising them as counter-examples for a refinement of the solution.

\subsection{Outline of the paper}
The paper is organised as follows: In Section \ref{lit}, a review of related work is provided. A brief overview of the kernel null-space technique and its regression-based formulation is presented in Section \ref{back}. The proposed robust kernel spectral regression approach is introduced in Section \ref{meth}. The results of an experimental evaluation of the proposed methodology are presented and discussed in Section \ref{ev}. Finally, conclusions are drawn in Section \ref{conc}.

\section{Related Work}
\label{lit}
Similar to the kernel null-space methodology, there exist other approaches operating on a Rayleigh quotient criterion. As an instance of the unsupervised methods in this category, the work in \cite{HOFFMANN2007863} utilises kernel PCA for novelty detection where the reconstruction residual of a test sample with respect to the inferred subspace is considered as a novelty measure. Other work \cite{GueSchVis07} proposes a strategy to improve the convergence of the kernel algorithm based on an iterative kernel PCA. A robustified PCA to deal with outliers in the training set is proposed in \cite{4522554}. A one-class kernel Fisher discriminant classifier is presented in \cite{NIPS2004_2656,6795824} which is based on the idea of separating the data from their negatively replicated counterparts and involves an eigenvalue decomposition of the kernel matrix. In this approach, once the data are mapped onto the feature space, a Mahalanobis distance to the mean of the fitted Gaussian is used as a test statistic. As indicated in \cite{6795824}, for kernel maps projecting input data into a higher-dimensional space, the Gaussianity assumption for the target data may not hold in general. An increasing deviation from normality may then lead to unreliable results of the method presented in \cite{NIPS2004_2656,6795824}. Other work in \cite{6619277} proposed a Fisher-based null space method where all training samples of one class are projected onto a single point. The proposed method treats multiple known classes jointly and  detects novel instances with respect to the set of classes using a single model operating on a joint subspace where the training samples of all known classes are presumed to have zero variance. In a follow-up work \cite{7045967}, it is proposed to incorporate locality in the null space approach of \cite{6619277} by considering only the most similar patterns to a query sample. In \cite{8099922}, an incremental version of the method in \cite{6619277} is proposed to increase computational efficiency. A generalised Rayleigh quotient specifically designed for outlier detection is presented in \cite{6857384,DUFRENOIS201696} where the method tries to find an optimal hyperplane which is closest to the target data and farthest from the outliers utilising two scatter matrices corresponding to the outliers and target data. In \cite{DUFRENOIS201696}, the generalised eigenvalue problem is replaced by an approximate conjugate gradient solution to moderate the computational cost of the method in \cite{6857384}. A later study \cite{7727608} tries to address limitations of the method in \cite{6857384,DUFRENOIS201696} in terms of the availability of outlier samples and difference in the densities of target and non-target observations via a null-space approach. While the majority of existing work on one-class classification using a Rayleigh quotient formulation requires computationally intensive eigen-decomposition of large matrices, the work in \cite{DBLP:journals/corr/abs-1807-01085}, presents a one-class approach which replaces costly eigen-analysis computations by a regression-based formulation \cite{SRKDA}. Among other unsupervised novelty detection techniques, the DPCP approach \cite{NIPS2018_7486} learns a linear subspace from data corrupted by outliers based on a non-convex $l_1$ optimisation problem. It is shown that DPCP can tolerate as many outliers as the square of the number of inliers, thus improving upon other robust PCA methods. Other method known as Outlier Pursuite (OP) is an efficient convex optimisation-based algorithm \cite{6126034} to perform a robust principal component analysis that under mild assumptions on the uncorrupted points recovers the exact optimal low-dimensional subspace and identifies the corrupted points. The FMS method is a non-convex robust subspace recovery approach \cite{doi:10.1093/imaiai/iax012}, designed to be least affected by corruptions in the training set and has been demonstrated to converge to a close vicinity of the correct subspace within few iterations. A leading unsupervised technique among others is that of SRO \cite{8099943} which obtains a weighted directed graph, defines a Markov Chain via self-representation, and identifies outliers via random walks.

\section{Background}
\label{back}
A brief overview of the kernel Fisher null-space approach \cite{DBLP:journals/corr/abs-1807-01085,6619277,8099922} and its regression-based formulation \cite{DBLP:journals/corr/abs-1807-01085} is provided next.
\subsection{Null-space Fisher analysis}
\label{NSKSR}
In statistical pattern classification, a widely used criterion for classification is that of Fisher where one tries to infer a projection function from the input space onto a subspace such that the within-class scatter of the data associated with each class is minimised while maximising the between-class scatter. More specifically, in a Fisher classifier, maximisers of the criterion function $\mathbf{J(\varphi)}$ are sought:
\begin{eqnarray}
 \operatorname*{arg\,max}_{\varphi}\mathbf{J(\varphi)}=\operatorname*{arg\,max}_{\varphi}\frac{\mathbf{\varphi^\top S_b \varphi}}{\mathbf{\varphi^\top S_w \varphi}}
\label{null}
\end{eqnarray}
\noindent where $\mathbf{S_b}$ denotes the between-class scatter matrix, $\mathbf{S_w}$ denotes the within-class scatter matrix and $\mathbf{\varphi}$ is the basis defining the subspace. In a one-class classification problem and in the absence of non-target training observations, the origin may be used as a counter-example. With regards to the Fisher criterion, a theoretically optimal projection is the one yielding a zero within-class scatter while providing a positive between-class scatter, referred to as a \textit{null} projection \cite{DBLP:journals/corr/abs-1807-01085,6619277,8099922}. Thus, in a null-space Fisher classifier:
\begin{eqnarray}
\nonumber \mathbf{\varphi^\top S_w \varphi} = 0\\
\mathbf{\varphi^\top S_b \varphi} > 0
\label{nullF}
\end{eqnarray}
A null projection function corresponds to the optimum of $\mathbf{J(\varphi)}$ in Eq. \ref{null} and thus provides the best separability with respect to the Fisher criterion. It can be shown that one may compute at most $C-1$ null projection directions, with $C$ being the number of classes. In a one-class formulation, since target observations are assumed to form a single class while (hypothetical) non-target samples correspond to a second class, only a single optimiser for Eq. \ref{null}, given as the eigenvector corresponding to the largest eigenvalue of the following eigen-problem exists:
\begin{eqnarray}
\mathbf{S_b\varphi}=\lambda \mathbf{S_w\varphi}
\label{geig}
\end{eqnarray}
Once the null projection direction is determined, the projection of a sample $x$ onto the null-space (hereafter referred to as response) is found as
\begin{eqnarray}
\mathbf{y} = \mathbf{\varphi}^\top \mathbf{x}
\label{nullproj}
\end{eqnarray}
In order to handle data with inherently non-linear structure, non-linear (kernel) extensions of this methodology are proposed \cite{DBLP:journals/corr/abs-1807-01085,6619277,8099922}. In kernel methods, a kernel function is utilised to implicitly project the data into a high dimensional space, known as the reproducing kernel Hilbert space (RKHS) in an attempt to make the data more easily separable in this new space. These methods typically require eigen-decompositions of dense matrices.

\subsection{One-Class Kernel Spectral Regression}
As eigen-computations associated with the kernel null-space technique are computationally demanding, an alternative approach based on spectral regression (called one-class kernel spectral regression, a.k.a. OC-KSR) was proposed in \cite{DBLP:journals/corr/abs-1807-01085}. The OC-KSR method is based on two principles: (1) in order for the within-class scatter to be zero, observations corresponding to a specific class are required to be mapped onto the same point in an optimal feature subspace, i.e. the responses (elements of $\mathbf{y}$) corresponding to samples of a particular class must be equal; and (2) in a RKHS, the function realising the above projection from the input space onto a feature subspace can be represented in terms of real numbers $\alpha_{i}$'s and a positive semi-definite kernel function $\kappa(.,.)$ as
\begin{eqnarray}
f(.)=\sum_{i=1}^{n}\alpha_i \kappa(.,x_i)
\label{projfunc}
\end{eqnarray}
\noindent where $x_i$'s denote training observations. In the OC-KSR approach \cite{DBLP:journals/corr/abs-1807-01085}, finding the optimal coefficients $\alpha_i$'s is then posed as a regression problem:
\begin{eqnarray}
f(.)^{opt}=\operatorname*{arg\,min}_{f(.)}\left \{\sum_{i=1}^{n}(f(x_i)-y_i)^2+\delta \Vert f \Vert^2_{K}\right \}
\end{eqnarray}
or equivalently as
\begin{eqnarray}
\boldsymbol{\alpha}^{opt}=\operatorname*{arg\,min}_{\boldsymbol{\alpha}}\left \{\Vert \mathbf{K}\boldsymbol{\alpha}-\mathbf{y} \Vert^2+\delta \boldsymbol{\alpha}^\top \mathbf{K} \boldsymbol{\alpha} \right\}
\label{reg}
\end{eqnarray}
\noindent where $\delta$ is a regularisation parameter and $\Vert . \Vert^2_K$ denotes the norm in the RKHS while $\mathbf{K}$ stands for the kernel matrix. The optimal solution $\boldsymbol{\alpha}$ to the problem above satisfies
\begin{eqnarray}
(\mathbf{K}+\delta \mathbf{I_n}) \boldsymbol{\alpha} = \mathbf{y}
\label{alphaeq}
\end{eqnarray}
\noindent where $\mathbf{I_n}$ denotes an identity matrix of size $n$. In the OC-KSR method of \cite{DBLP:journals/corr/abs-1807-01085}, the kernel function is an RBF and thus the kernel matrix is positive definite. As the kernel matrix is invertible, no regularisation is imposed (i.e. the regularisation parameter $\delta$ is set to zero) on $\boldsymbol\alpha$. Based on the availability of training data, two cases are considered in the OC-KSR method: 1-if only positive training samples are available, the response vector $\mathbf{y}$ is shown to be $\mathbf{y}=(\overbrace{1,\dots,1}^{n})^{\top}$. When both positive and negative training observations are available, the response vector $\mathbf{y}$ is given as $\mathbf{y}=(\overbrace{1,\dots,1}^{n-n_o},\overbrace{0,\dots,0}^{n_0})^{\top}$ (up to a scale factor) where $n_0$ denotes the number of negative training samples. Given $\mathbf{y}$, solving Eq. \ref{alphaeq} for $\boldsymbol{\alpha}$ is performed efficiently using the Cholesky factorisation via the Sherman's march algorithm \cite{matrixdecom}. Once $\boldsymbol{\alpha}$ is determined, a test sample may be projected onto the null feature subspace using Eq. \ref{projfunc}. In the decision making stage, the (Euclidean) distance between the projection of a test sample and that of the target class is used as a dissimilarity criterion.

\section{Robust One-Class Kernel Spectral Regression}
\label{meth}
In this section, the proposed approach to build a robust one-class classifier based on the null-space kernel methodology in presence of contaminations in the training set is presented. In the proposed method, starting from an initial assumption regarding observation labels, label confidences are updated iteratively and soft labels are assigned to the training observations reflecting their degree of normality. As a result, although the proposed method is designed to detect novelties with respect to a given set of training samples, nevertheless, it can be directly used for observation ranking purposes in an unsupervised setting as will be discussed in the subsequent sections.

As noted earlier, the proposed approach is based on a regularised regression formulation, i.e.
\begin{eqnarray}
P(\boldsymbol{\alpha},\mathbf{y})=\Vert\mathbf{K}\boldsymbol{\alpha}-\mathbf{y}\Vert^2+\mathcal{R}(\boldsymbol{\alpha})
\end{eqnarray}
where $\mathbf{y}$ is the expected responses (labels) for observations and $\mathcal{R}(\boldsymbol{\alpha})$ encodes a desired regularisation on the solution $\boldsymbol{\alpha}$.
The regularisation, in general, may serve different purposes. First, when the number of variables exceeds the number of observations, the least-squares problem is ill-posed and it is therefore impossible to solve the associated optimisation problem as infinitely many solutions exist. Regularisation in this case allows the introduction of additional constraints that help to uniquely determine the solution. The second case where regularisation may be deployed corresponds to the case where the number of variables does not exceed the number of samples, but the model learned suffers from poor generalisation capability. In such cases, regularisation is used to improve the generalisation performance of the model by constraining it during the training phase. The regularisation term thus imposes a limitation on the function space available by introducing a penalty to discourage certain regions of the function space. In the current work, given sparse and noisy samples of a function $f(.)$ from a corrupted data set, regularisation constraints the function and maintains a trade-off between data fidelity and some constraint on the solution function. The imposed constraint may, for example, enforce the solution to be sparse or to reflect other prior knowledge about the solution such as constraining its norm in the corresponding space. In such cases, regularisation methods typically correspond to priors on the solution to a least squares problem.

Following a regularised regression formulation, the proposed approach to handle contaminations in the training set is to optimise the objective function $P$ not only with respect to $\boldsymbol{\alpha}$ but also with respect to $\mathbf{y}$. While optimisation with respect to $\boldsymbol \alpha$ is the standard approach to specify the parameters characterising the projection function given by Eq. \ref{projfunc}, optimisation with respect to $\mathbf{y}$ reflects (and compensates for) the absence of prior knowledge regarding conformity of individual observations to the model. This is fundamentally different from the ordinary OC-KSR method \cite{DBLP:journals/corr/abs-1807-01085} where $\mathbf{y}$ is fixed and $\boldsymbol{\alpha}$ optimisers of $P$ are sought. Moreover, the method in \cite{DBLP:journals/corr/abs-1807-01085} does not impose a regularisation on the model when deriving the final solution (i.e. $\delta$ is set to $0$). In essence, given an initial guess regarding object labels (i.e. the responses $\mathbf{y}$), the proposed approach updates label confidences and derives soft labels for all training samples while at the same time optimising the objective function with respect to $\boldsymbol{\alpha}$. This is realised via a block coordinate descent minimisation approach alternating between minimising $P$ with respect to $\boldsymbol{\alpha}$ and minimising it with respect to $\mathbf{y}$. Optimising $P$ with respect to $\mathbf{y}$ can be realised by setting its partial derivative with respect to $\mathbf{y}$ equal to zero, i.e.
\begin{eqnarray}
\frac{\partial P}{\partial\mathbf{y}}=-2(\mathbf{K}\boldsymbol{\alpha}-\mathbf{y})=0
\end{eqnarray}
which gives $\mathbf{y}=\mathbf{K}\boldsymbol{\alpha}$. As such, the generic scheme of the proposed approach can be summarised as Algorithm \ref{generic}. It can be shown that the alternating minimisation scheme of Algorithm \ref{generic} is convergent. This can be readily confirmed by denoting $\boldsymbol{\alpha}_{t+1}=\operatorname*{arg\,min}_{\Vert\boldsymbol{\alpha}\Vert=1} P(\boldsymbol{\alpha},\mathbf{y}_t)$ and $\mathbf{y}_{t+1}=\operatorname*{arg\,min}_{\mathbf{y}} P(\boldsymbol{\alpha}_{t+1},\mathbf{y})$ for iteration $t$. Consequently, one has
\begin{eqnarray}
P(\boldsymbol{\alpha}_{t},\mathbf{y}_t)\geq P(\boldsymbol{\alpha}_{t+1},\mathbf{y}_t)\geq P(\boldsymbol{\alpha}_{t+1},\mathbf{y}_{t+1})
\end{eqnarray}
The regularisation types considered in this work are those of Tikhonov and sparsity. As a result, the objective function $P(\boldsymbol{\alpha},\mathbf{y})$ is bounded from below and thus the non-increasing sequence $\lim_{t \to \infty} P(\boldsymbol{\alpha}_{t},\mathbf{y}_t)$ is convergent.
\begin{algorithm}[t]
\footnotesize
\caption{The generic scheme of the proposed approach}\label{generic}
\begin{algorithmic}[1]
\Repeat
\State $\boldsymbol{\alpha}=\operatorname*{arg\,min}_{\Vert\boldsymbol{\alpha}\Vert=1} P(\boldsymbol{\alpha},\mathbf{y})$
\State $\mathbf{y}=\mathbf{K}\boldsymbol{\alpha}$
\Until{convergence}
\end{algorithmic}
\normalsize
\end{algorithm}

Solving the minimisation problem $\boldsymbol{\alpha}^{opt}=\operatorname*{arg\,min}_{\Vert\boldsymbol{\alpha}\Vert=1} P(\boldsymbol{\alpha},\mathbf{y})$ is dependent upon the specific regularisation imposed on the solution. Although other possibilities exist, in this work, two commonly used regularisation schemes of Tikhonov and sparsity are considered for $\mathcal{R}(\boldsymbol{\alpha})$, discussed next.
\subsection{Tikhonov regularisation}
In the case of a Tikhonov regularisation (also known as ridge regression), $\mathcal{R}(\boldsymbol{\alpha})=\delta \boldsymbol{\alpha}^\top \mathbf{K} \boldsymbol{\alpha}$ and the objective function $P(\boldsymbol{\alpha},\mathbf{y})$ would be
\begin{eqnarray}
P(\boldsymbol{\alpha},\mathbf{y})=\Vert \mathbf{K}\boldsymbol{\alpha}-\mathbf{y} \Vert^2+\delta \boldsymbol{\alpha}^\top \mathbf{K} \boldsymbol{\alpha}
\end{eqnarray}
In this case, the problem $\boldsymbol{\alpha}^{opt}=\operatorname*{arg\,min}_{\Vert\boldsymbol{\alpha}\Vert=1} P(\boldsymbol{\alpha},\mathbf{y})$ can be solved by setting the derivative of $P(\boldsymbol{\alpha},\mathbf{y})$ with respect to $\boldsymbol{\alpha}$ to zero, i.e.
\begin{eqnarray}
\frac{\partial P}{\partial\boldsymbol{\alpha}}=2\mathbf{K}(\mathbf{K}\boldsymbol{\alpha}-\mathbf{y})+2\delta \mathbf{K}\boldsymbol{\alpha}=0
\end{eqnarray}
which gives
\begin{eqnarray}
\boldsymbol{\alpha}^{opt}=(\mathbf{K}+\delta\mathbf{I}_n)^{-1}\mathbf{y}
\end{eqnarray}
Combining the two equations for $\boldsymbol{\alpha}$ and $\mathbf{y}$, the proposed approach based on a Tikhonov regularisation is given as Algorithm \ref{Tik}.
\begin{algorithm}[t]
\footnotesize
\caption{The proposed approach based on Tikhonov regularisation}\label{Tik}
\begin{algorithmic}[1]
\Repeat
\State $\boldsymbol{\alpha}=(\mathbf{K}+\delta\mathbf{I}_n)^{-1}\mathbf{y}$
\State $\boldsymbol{\alpha}=\boldsymbol{\alpha}/\Vert \boldsymbol{\alpha}\Vert$
\State $\mathbf{y}=\mathbf{K}\boldsymbol{\alpha}$
\Until{convergence}
\end{algorithmic}
\normalsize
\end{algorithm}
In a way, Tikhonov regularisation favours models that provide predictions that are as smooth functions of the data as
possible. In other words, such a regularisation scheme penalises larger values taken by the solution coefficients, thereby producing a more parsimonious solution incorporating a set of coefficients with smaller variance which is particularly advantageous when making inference in a noisy data set.
\subsubsection{Optimal regularisation parameter}
Algorithm \ref{Tik} provides a procedure to find the optimal parameters $\boldsymbol{\alpha}$ and $\mathbf{y}$. Yet, it does not specify how to choose the Tikhonov regularisation parameter $\delta$.
In order to infer the optimal Tikhonov regularisation parameter, the one-class learning problem in presence of outliers is posed as a sensitivity analysis problem in this work. In this respect, the optimal Tikhonov regularisation parameter is derived so as to minimise the sensitivity of the solution $\boldsymbol\alpha$ with respect to contaminations in the training set. For this purpose, let us assume a set of contaminated observations $X = \{x_1, \dots, x_{n} \}$ for which the \textit{true} labels are recorded in vector $\mathbf{y}$. Apparently one is not informed of the true labels in a contaminated data set. Instead, an initial guess for object labels may be assumed. In the absence of any prior knowledge, the initial assumption for all observations (recorded in vector $\mathbf{y}^\prime$) may be that of being \textit{target} samples, i.e. $\mathbf{y}^\prime=\{\overbrace{1,\dots,1}^{n}\}$. $\mathbf{y}^\prime$ corresponds to the noisy assumption of labels which deviates from the true labels by $\Delta\mathbf{y}$, i.e. $\mathbf{y}^\prime=\mathbf{y}+\Delta \mathbf{y}$. Using the assumed noisy labels $\mathbf{y}^\prime$, the weight vector $\boldsymbol{\alpha}^\prime$ for the Tikhonov regularised problem is given as $\boldsymbol{\alpha}^\prime=(\mathbf{K}+\delta I)^{-1}\mathbf{y}^\prime$. In practice, it is desirable to have $\boldsymbol{\alpha}^\prime$ as close as possible to the true $\boldsymbol{\alpha}$ (derived based on $\mathbf{y}$) so that the responses obtained as $\mathbf{K}\boldsymbol{\alpha}^\prime$ are as close as to true labels $\mathbf{y}=\mathbf{K}\boldsymbol{\alpha}$. This is essentially a sensitivity analysis problem where one is given a perturbed vector $\mathbf{y}^\prime=\mathbf{y}+\Delta\mathbf{y}$ and the goal is to find a solution $\boldsymbol{\alpha}^{\prime}$ with minimal difference from the ideal solution $\boldsymbol{\alpha}$. Assuming $\Vert \boldsymbol{\alpha}^\prime\Vert > 0$, the sensitivity of the solution with respect to perturbations in object labels is defined as
\begin{eqnarray}
S=\frac{\Vert \boldsymbol{\alpha}^\prime -\boldsymbol{\alpha} \Vert}{\Vert\boldsymbol{\alpha}^\prime \Vert}
\end{eqnarray}
Apparently, minimising the sensitivity of the solution would maximise the similarity between the true labels $\mathbf{y}$ and the inferred responses $\mathbf{y}^\prime$. The sensitivity of the regression solution in the Hilbert space using a Tikhonov regularisation in a general setting is studied in \cite{10.1007/978-3-540-87536-9_23} and summarised in the Theorem below and the following Corollary.
\begin{thm}
Let $\kappa(.,.)$ be a kernel function and let the kernel matrix $\mathbf{K}$ corresponding to a set of observations $X$ be positive definite. Then the optimal value of parameter $\delta > 0$ with respect to the sensitivity of the regularised solution of the problem $E_Z+\delta \Vert . \Vert^2_{K}$ is
\begin{eqnarray}
\delta_{opt}=\frac{\lambda_{min}(c(\mathbf{K})-\frac{c(\mathbf{K})+1}{2\sqrt{c(\mathbf{K})}})}{\frac{c(\mathbf{K})+1}{2\sqrt{c(\mathbf{K})}}-1}
\end{eqnarray}
where $E_Z=\frac{1}{n}\Vert \mathbf{K} \boldsymbol{\alpha}-\mathbf{y}\Vert^2$
\end{thm}
\noindent $\lambda_{min}$ denotes the smallest eigenvalue of the kernel matrix $\mathbf{K}$ and $c(.)$ stands for the condition number of a matrix, defined as the ratio of the largest eigenvalue to the smallest one.

\begin{cor}
In addition to the requirements of the Theorem above, if the kernel matrix is normalised, then the optimal value of parameter $\delta > 0$ with respect to the sensitivity of the regularised solution of the problem $E_Z+\delta \Vert . \Vert^2_{K}$ is given as
\begin{eqnarray}
\delta_{opt}=\frac{1}{1+\lambda_{min}}-\frac{\lambda_{min}(2-\sqrt{\lambda_{min}})}{2}
\label{optp}
\end{eqnarray}
\end{cor}
For a proof of the theorem and the corresponding corollary cf. \cite{10.1007/978-3-540-87536-9_23}.

As the kernel function used in this work is that of a radial basis function, the kernel matrix would be positive definite. Consequently, for a minimum sensitivity solution in the proposed approach, one may follow Algorithm \ref{Tik}, setting the regularisation parameter based on Eq. \ref{optp}.

\subsection{Sparse regularisation}
In addition to the widely used Tikhonov regularisation, other regularisation approaches encouraging sparseness of the solution have a rich history as a guideline for inference. The underlying motivation for seeking a sparse characterisation is the desire to provide the simplest possible explanation of an observation as a linear combination of as few as possible atoms from a given dictionary. One of the most celebrated instantiations of sparseness, the principle of minimum description length in model selection requires that within a pool of models, the model that yields the most compact representation should be preferred for decision making. Such sparse methods select a small subset of relevant atoms to characterise the solution. A sensible sparsity constraint to impose on the solution of the kernel spectral regression approach is the $l_{0}$-norm, defined as the number of non-zero elements in the solution. However, solving an $l_{0}$-regularised learning problem has been demonstrated to be NP-hard. The $l_{1}$-norm is shown to induce sparsity and can be used to approximate the optimal $l_{0}$-norm via convex relaxation. A least squares problem in the presence of an $l_1$ regularisation term is known as Lasso in statistics and basis pursuit in signal processing. Sparse $l_{1}$-norm models allow for scalable algorithms that can handle problems with a large number of parameters. Encouraging the solution of the kernel null-space approach to be sparse can be conveniently performed by enforcing an $l_{1}$-regulariser, i.e. $\mathcal{R}=\delta \sum_{i=1}^n |\alpha_i|$. Consequently, the objective function $P(\boldsymbol{\alpha},\mathbf{y})$ in this case would be 
\begin{eqnarray}
P(\boldsymbol{\alpha},\mathbf{y}) = \Vert \mathbf{K}\boldsymbol{\alpha}-\mathbf{y} \Vert^2+\delta \sum_{i=1}^n |\alpha_i|
\label{lasso}
\end{eqnarray}
The degree of sparseness of a solution is controlled via parameter $\delta$. In a sparse formulation of the regression problem, each response $y_i$ is generated using only a few observations from the training set. In particular, if the solution $\boldsymbol{\alpha}$ is very sparse, a large number of observations would have no contribution to the final solution, the immediate implication of which is a reduction in the computational complexity of the algorithm in the test phase. A further and more important consequence of forming a sparse model, as also suggested by other studies \cite{4483511}, is that of classification performance where a more compact model could improve performance compared with its non-sparse counterpart, especially in presence of corrupted data.

Efficient solving of a lasso problem is a subject of ongoing and fast developing research. While efficiently solving the lasso problem is desirable, yet, as this stage of the method is presumed to be performed offline, it has no impact on the efficiency of the proposed approach in the test phase. The complexity of the method in the test phase is controlled by the degree of sparsity of the solution, i.e. $\boldsymbol{\alpha}$ and not by a procedure to derive such a solution. In this work, the Least Angel Regression (LARS) algorithm \cite{efron_2004} is used to find the optimal solution corresponding to $\boldsymbol{\alpha}=\operatorname*{arg\,min}_{\boldsymbol{\alpha}} \Vert \mathbf{K}\boldsymbol{\alpha}-\mathbf{y} \Vert^2+\delta \sum_{i=1}^n |\alpha_i|$. Using the LARS algorithm, solutions with all possible cardinalities on $\boldsymbol{\alpha}$ can be computed. For the lasso formulation, the proposed approach is given in Algorithm \ref{lasso1}.

\begin{algorithm}[t]
\footnotesize
\caption{The proposed approach based on Lasso regression}\label{lasso1}
\begin{algorithmic}[1]
\Repeat
\State $\boldsymbol{\alpha}=\operatorname*{arg\,min}_{\boldsymbol{\alpha}} \Vert \mathbf{K}\boldsymbol{\alpha}-\mathbf{y} \Vert^2+\delta \sum_{i=1}^n |\alpha_i|$
\State $\boldsymbol{\alpha}=\boldsymbol{\alpha}/\Vert \boldsymbol{\alpha}\Vert$
\State $\mathbf{y} = \mathbf{K}\boldsymbol{\alpha}$
\Until{convergence}
\end{algorithmic}
\normalsize
\end{algorithm}

\subsection{Known fraction of contaminations}
In the discussions thus far, it is assumed that the fraction of training data corresponding to contaminations in the training set is not known in advance. In case the number of contaminations in the training set (denoted as $n_0$) is known, this information can be incorporated into the learning algorithm. This is realised by ranking training observations according to their compatibility with the model and identifying $n_0$ least compatible observations as non-target samples which results in an splitting of the training set into a target set and a non-target set. For this purpose, at each iteration, the responses for all observations are sorted and the smallest $n_0$ elements of $\mathbf{y}$ are set to zero while others are set to one. This procedure essentially corresponds to updating the initial normality assumption  regarding observations which is expected to improve the performance. In this respect, the samples corresponding to the $n_0$ samples which are less likely to be positive observations would form a non-target set which are used for the refinement of the solution. The thus obtained generic scheme is given as Algorithm \ref{known_reg} where the $\texttt{SR}(.)$ routine corresponds to a sorting and updating of the elements of an argument vector as described above. The minimisation step $\boldsymbol{\alpha}=\operatorname*{arg\,min}_{\Vert\boldsymbol{\alpha}\Vert=1} P(\boldsymbol{\alpha},\mathbf{y})$ in Algorithm \ref{known_reg} is performed as discussed previously, depending on the specific type of regularisation employed.
\begin{algorithm}[t]
\footnotesize
\caption{The generic scheme of the proposed approach when the fraction of contaminations is known}\label{known_reg}
\begin{algorithmic}[1]
\Repeat
\State $\boldsymbol{\alpha}=\operatorname*{arg\,min}_{\Vert\boldsymbol{\alpha}\Vert=1} P(\boldsymbol{\alpha},\mathbf{y})$
\State $\mathbf{y} = \texttt{SR}(\mathbf{K}\boldsymbol{\alpha})$
\Until{convergence}
\end{algorithmic}
\normalsize
\end{algorithm}

\subsection{Decision strategy}
\label{decr}
Once the optimal projection parameter $\boldsymbol\alpha$ is inferred, the projection of a test sample onto the feature subspace is realised as per Eq. \ref{projfunc}. The projections of target samples in the proposed formulation are expected to lie at points closer to point 1 in the feature subspace while those corresponding to non-target samples are expected to lie at points closer to the origin. Consequently, the decision rule for a test sample $z$ is defined as
\begin{eqnarray}
f(z)=\sum_{i=1}^{n}\alpha_i \kappa(z,x_i) \geq \tau& \hspace{1cm} &\mbox{\textit{z is a target object}}\nonumber \\ 
f(z)=\sum_{i=1}^{n}\alpha_i \kappa(z,x_i)  < \tau& \hspace{1cm} &\mbox{\textit{z is an outlier}}
\label{dr}
\end{eqnarray}
where $\tau$ is a threshold for deciding normality.

\section{Experimental Evaluation}
\label{ev}
In this section, an experimental evaluation of the proposed approach along with a comparison to the state-of-the-art methods is presented. On each data set, the training data includes both positive and negative instances to simulate a contaminated training set.
\subsection{Data sets}
\subsubsection{Face} This data set is created to perform a toy experiment in face recognition. The data set contains face images of different individuals where the task is to recognise a subject among others. For each subject, a one-class classifier is built using the training data associated with the subject under consideration while all other subjects are assumed as outliers with respect to the built model. The experiment is repeated in turn for each subject in the dataset. The features used for face image representation are obtained via the frontal-pose PAM deep CNN model \cite{8255649} applied on the face bounding boxes. The data set is created out of the real-access videos of the Replay-Mobile dataset \cite{Costa-Pazo_BIOSIG2016_2016} which provides face bounding boxes. In this work, ten subjects are used to form the data set where each subject is represented using 30 positive training instances. The number of negative training observations for each subject, i.e. contaminations, is also 30 images selected randomly from subjects other than the subject under consideration. The number of positive and negative test samples are similarly set to 30 images each.
\subsubsection{MNIST} MNIST is a collection of $28\times 28$ pixel images of handwritten digits 0-9 \cite{726791}. In our experiments, a single digit is considered as the target digit while all others correspond to non-target observations. In the experiments on this data set, the number of positive and negative training instances are set to 50 images each. Similarly, 50 positive and 50 negative images are included in the test set. The target class is assumed to be digit '3' while all other digits represent anomalies with respect to the target class.
\subsubsection{Coil-100} The Coil-100 data set \cite{coil} contains 7,200 images of 100 different objects. Each object has 72 images taken at pose intervals of 5 degrees, with the images being of size $32\times 32$ pixels. In the experiments conducted on this data set, a single object is randomly chosen to be the target class while all others are considered as novelties. Raw pixel intensities are used as feature representations in this data set. The number of positive train and test instances for the target class is 36 each. Similarly, 36 negative train and 36 test observations are included in the experiments on this data set.

\subsection{Convergence behaviour}
\begin{figure}[t]
\centering
\includegraphics[scale=.13]{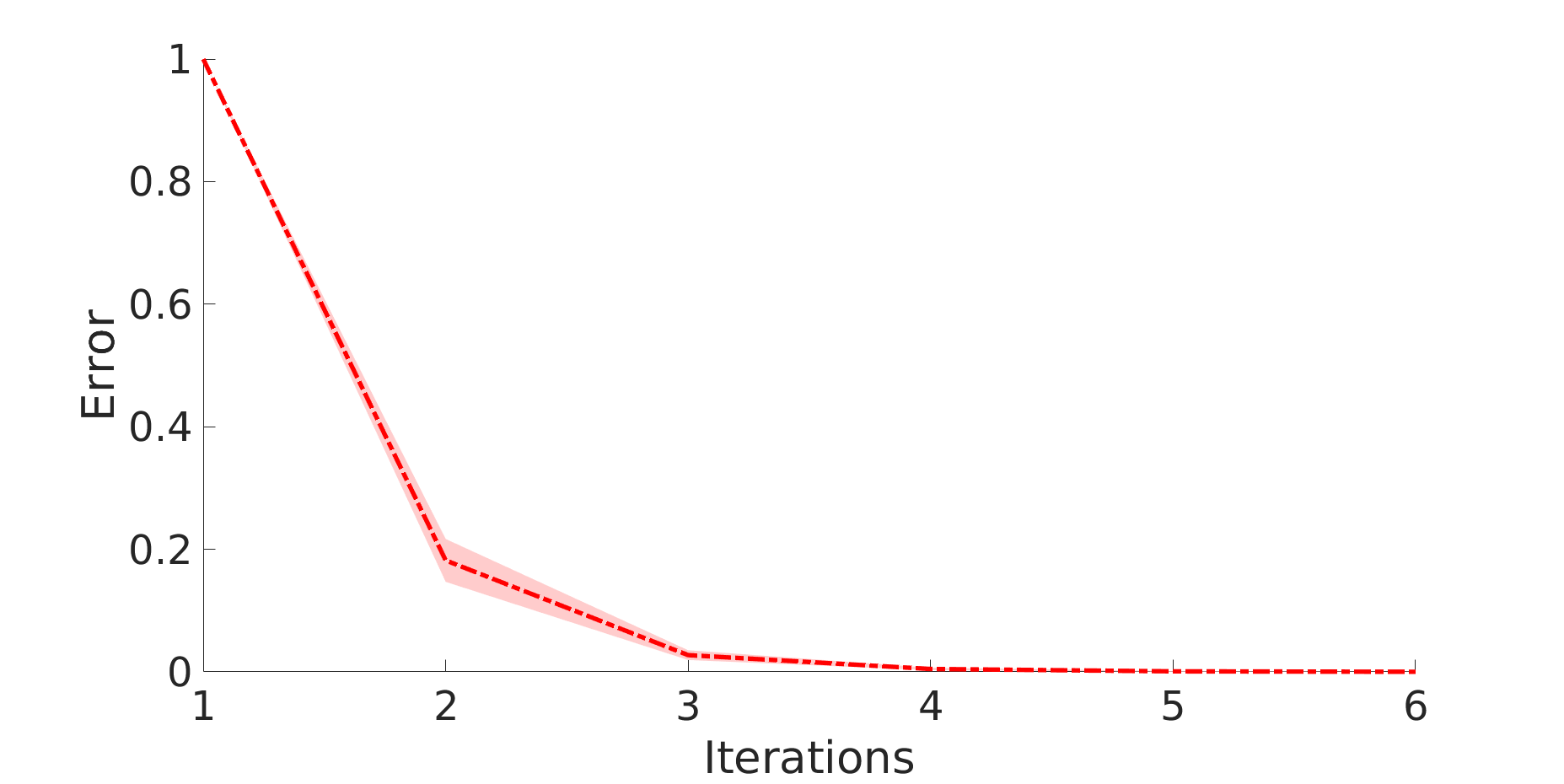}
\includegraphics[scale=.13]{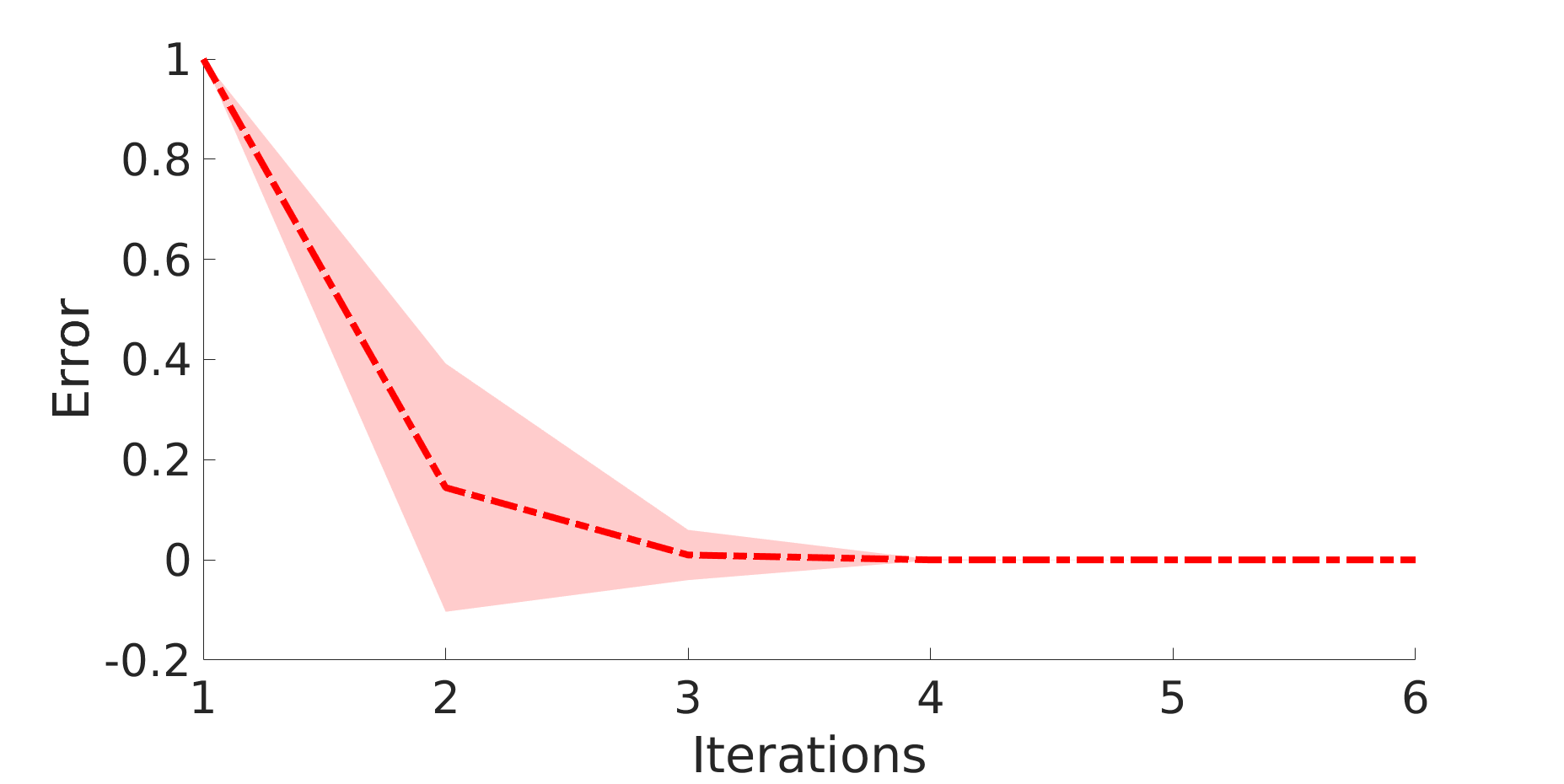}
\caption{Convergence behaviour (mean$±\pm$std) of the proposed iterative methods on the face data set; top: Tikhonov regularisation, bottom: Sparse regularisation}
\label{con_face}
\end{figure}

\begin{figure}[t]
\centering
\includegraphics[scale=.13]{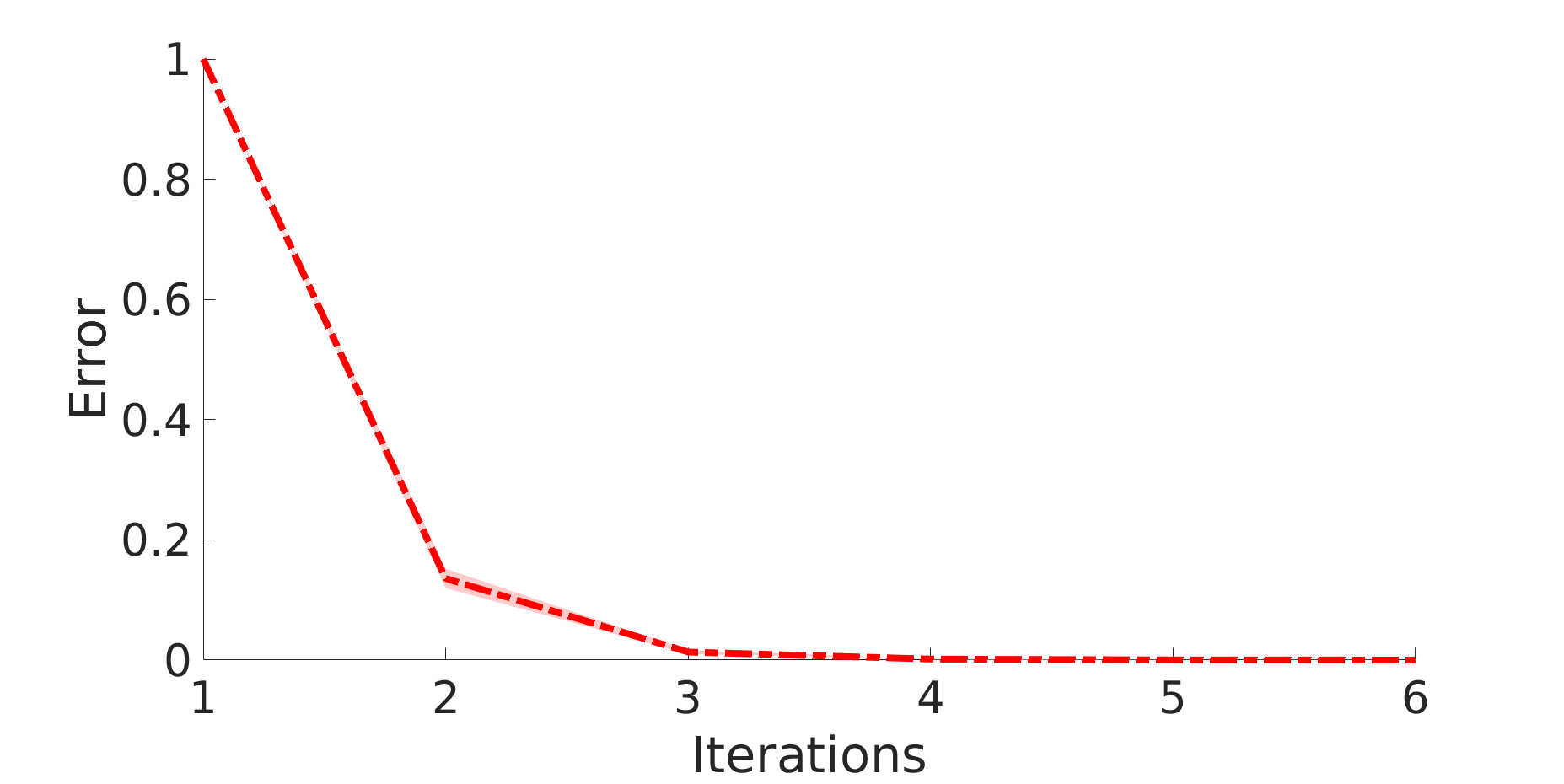}
\includegraphics[scale=.13]{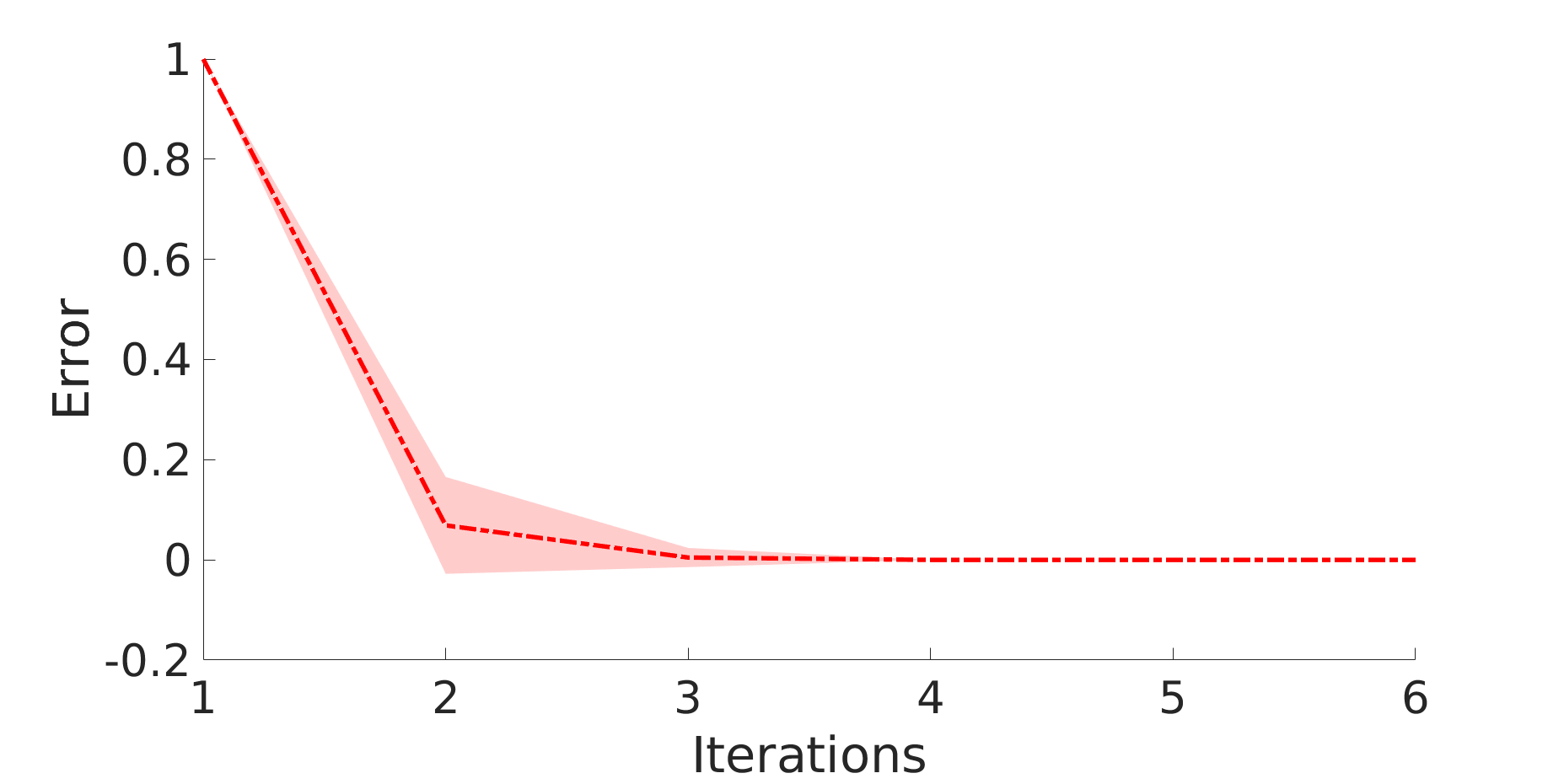}
\caption{Convergence behaviour (mean$±\pm$std) of the proposed iterative methods on the MNIST data set; top: Tikhonov regularisation, bottom: Sparse regularisation}
\label{con_mnist}
\end{figure}

\begin{figure}[t]
\centering
\includegraphics[scale=.13]{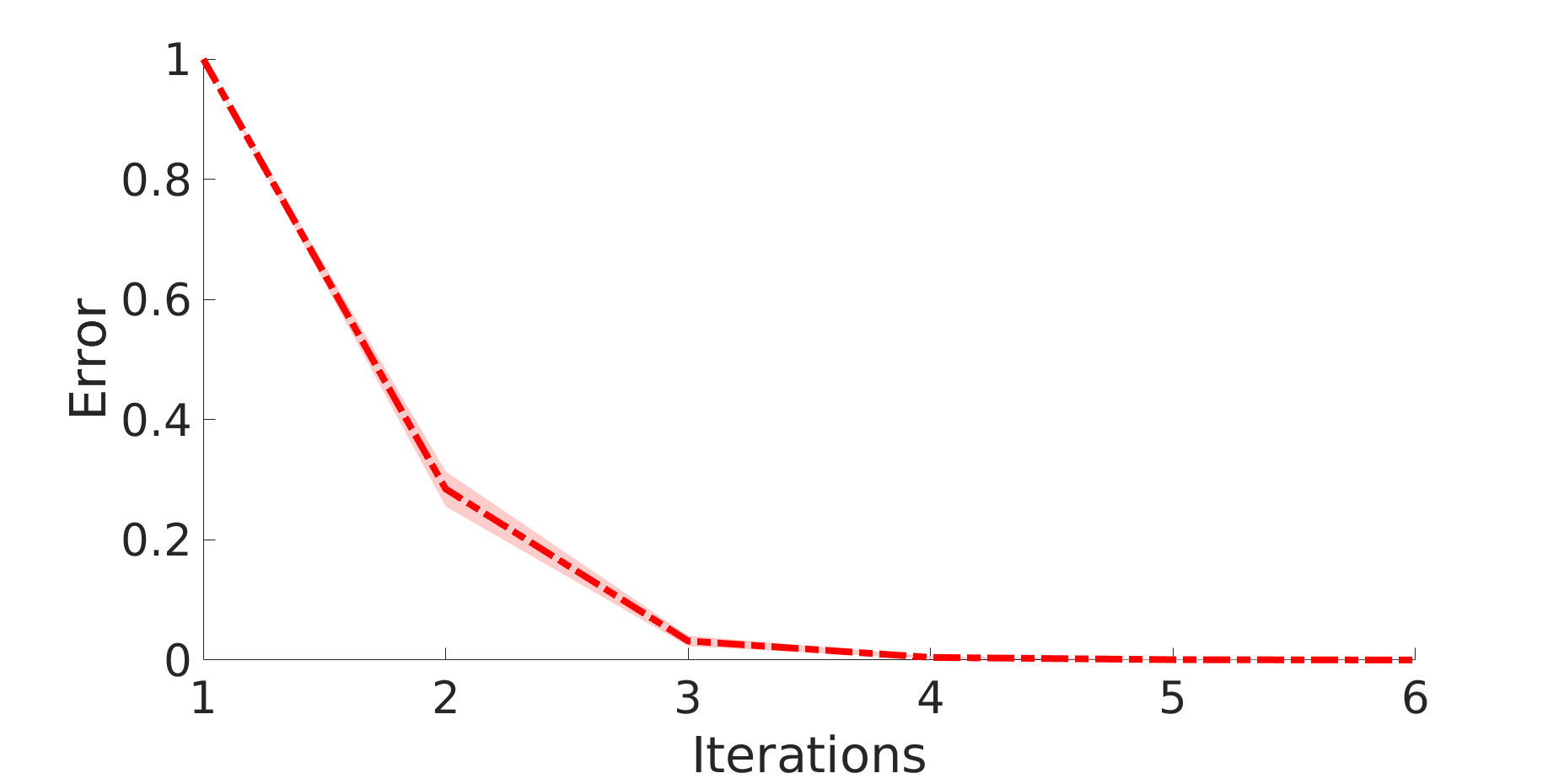}
\includegraphics[scale=.13]{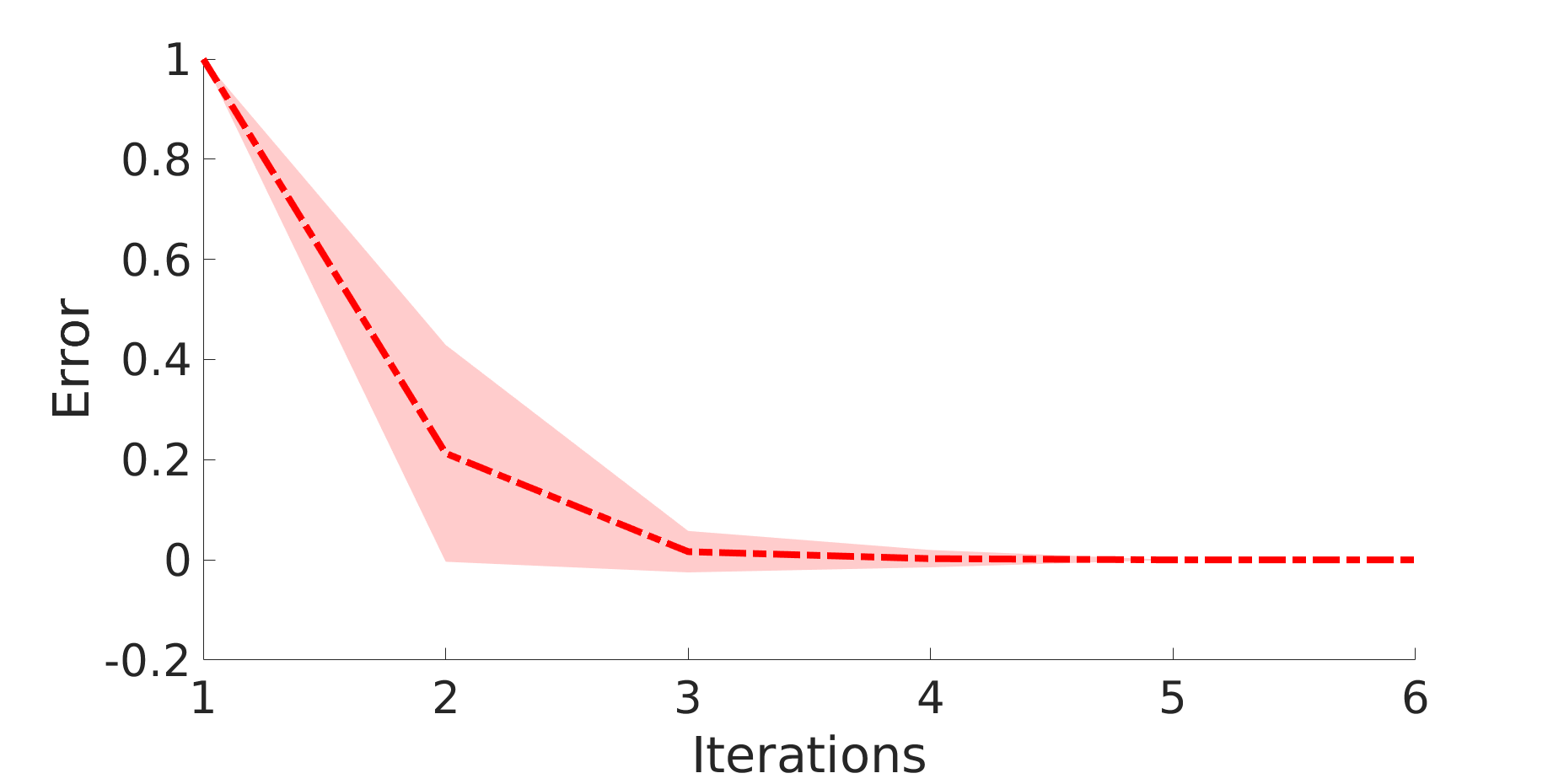}
\caption{Convergence behaviour (mean$±\pm$std) of the proposed iterative methods on the Coil-100 data set; top: Tikhonov regularisation, bottom:Sparse regularisation}
\label{con_coil}
\end{figure}

In this section, the convergence behaviour of the proposed block coordinate descent method for optimisation is analysed. For this purpose, the proposed models are trained on the face, MNIST and Coil-100 data sets and the error, defined as the norm of the deviation in $\boldsymbol\alpha$ between two consecutive iterations, vs. number of iterations of the corresponding algorithms are recorded over ten random splits of data into training and test sets. A zero norm deviation would be indicative of the convergence of the optimisation algorithm. The mean and std. (shaded regions in the figure) of the error vs. iterations are given in Fig. \ref{con_face}, \ref{con_mnist} and \ref{con_coil} for the face, MNIST and Coil-100 data sets, respectively. As can be observed from the figures, on all three data sets, the algorithm, on average, converges faster for the Tikhonov regularisation as compared with the sparse model. Nevertheless, regardless of the regularisation type imposed on the solution, convergence is quite fast where the methods converge in just a few number of iterations (on average in as few as five iterations).

\subsection{The effect of regularisation parameter}
\begin{figure}[t]
\centering
\includegraphics[scale=.17]{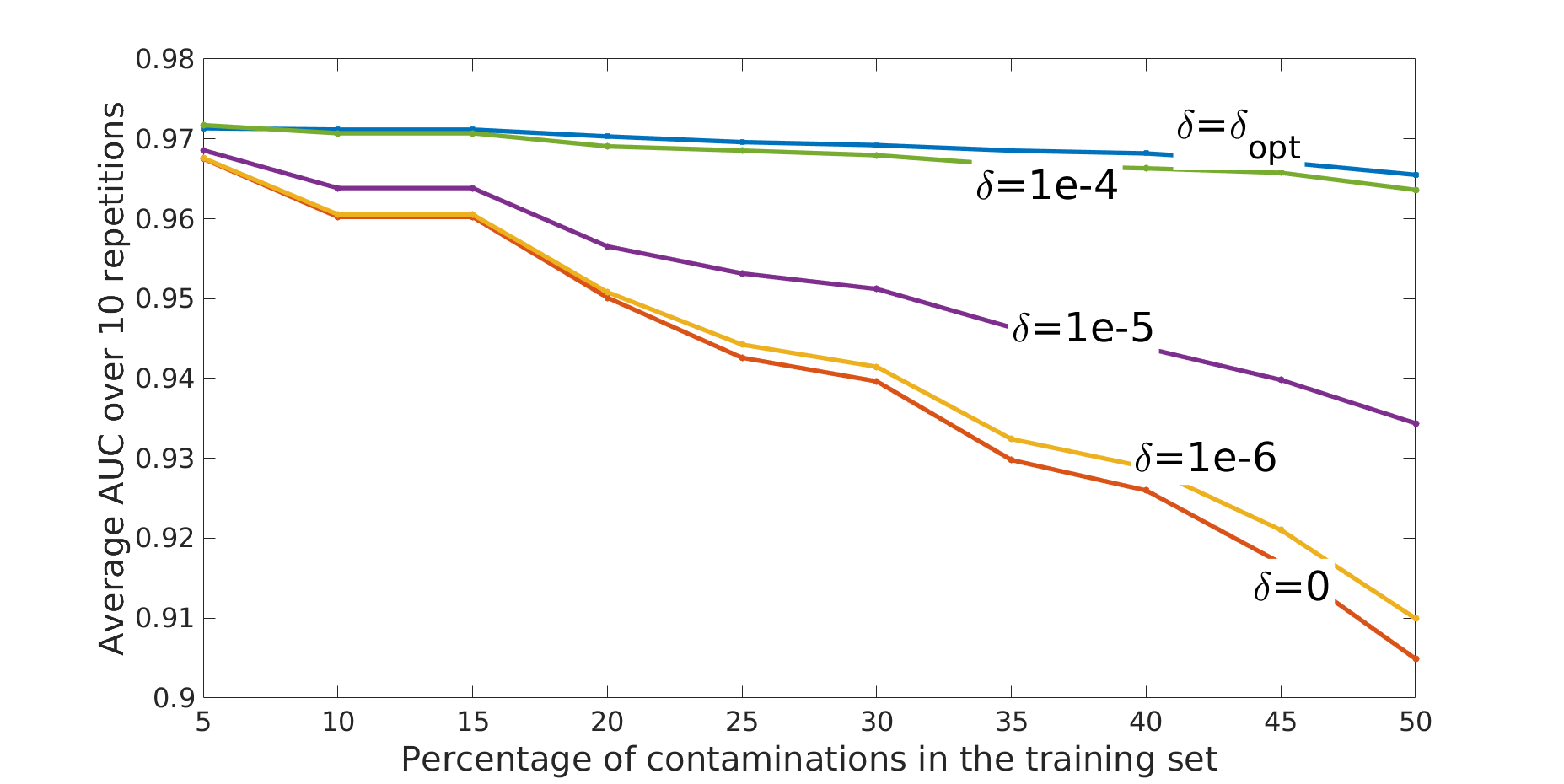}
\includegraphics[scale=.17]{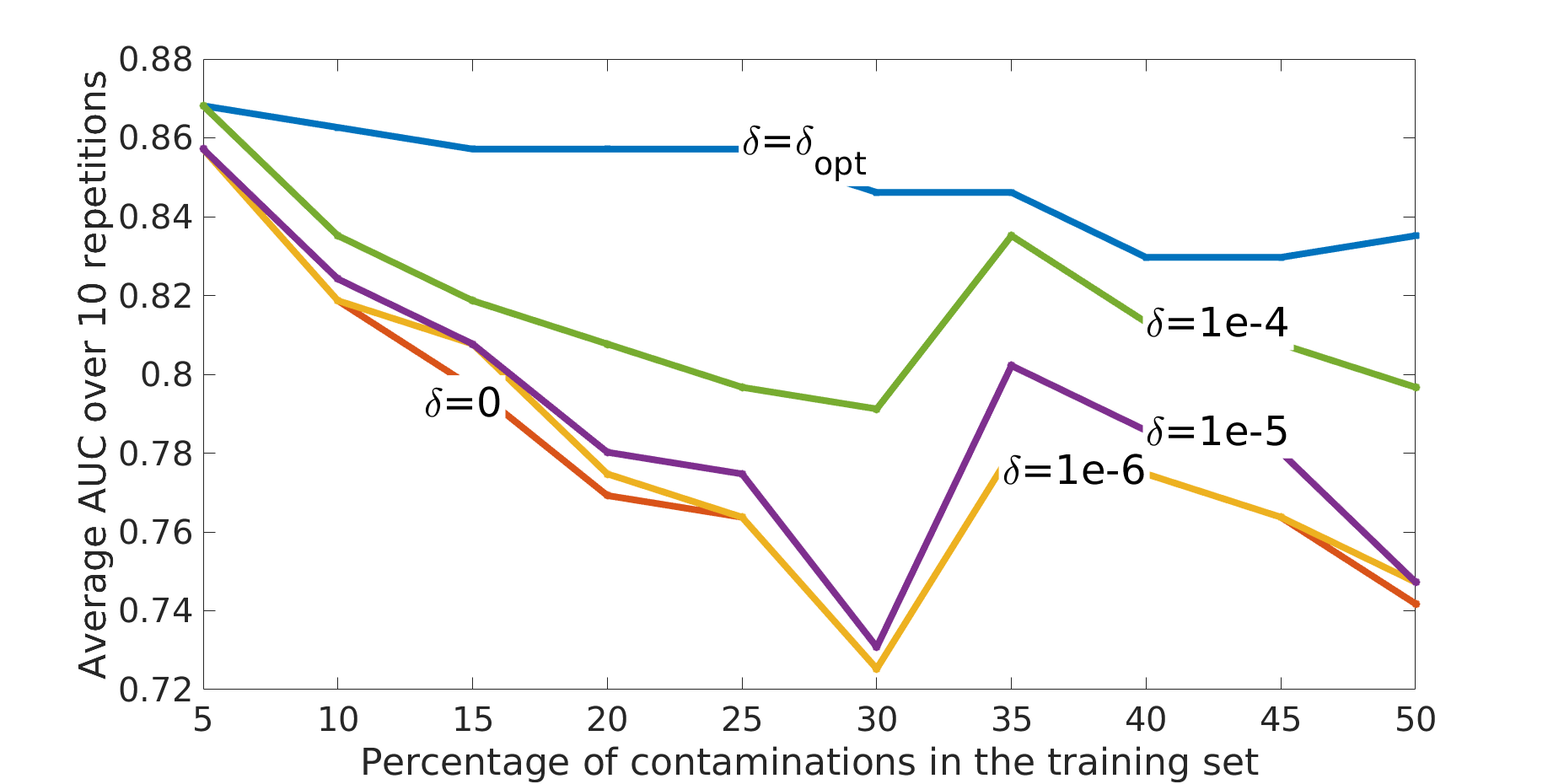}
\includegraphics[scale=.17]{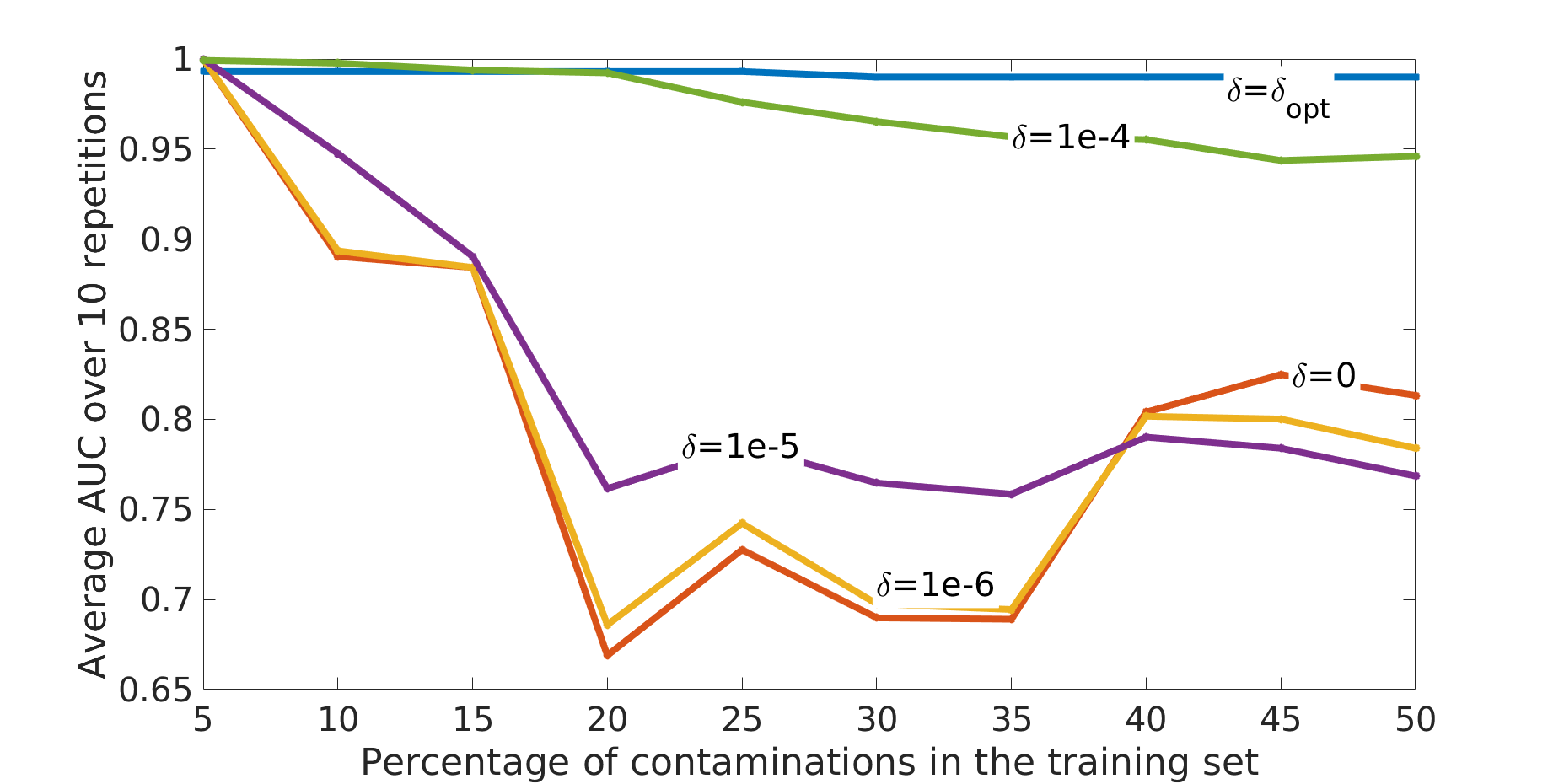}
\caption{The effect of varying the Tikhonov regularisation on the performance of the proposed approach. From top to bottom: face data set, MNIST data set and the Coil-100 data set}
\label{var1}
\end{figure}

\begin{figure}[t]
\centering
\includegraphics[scale=.17]{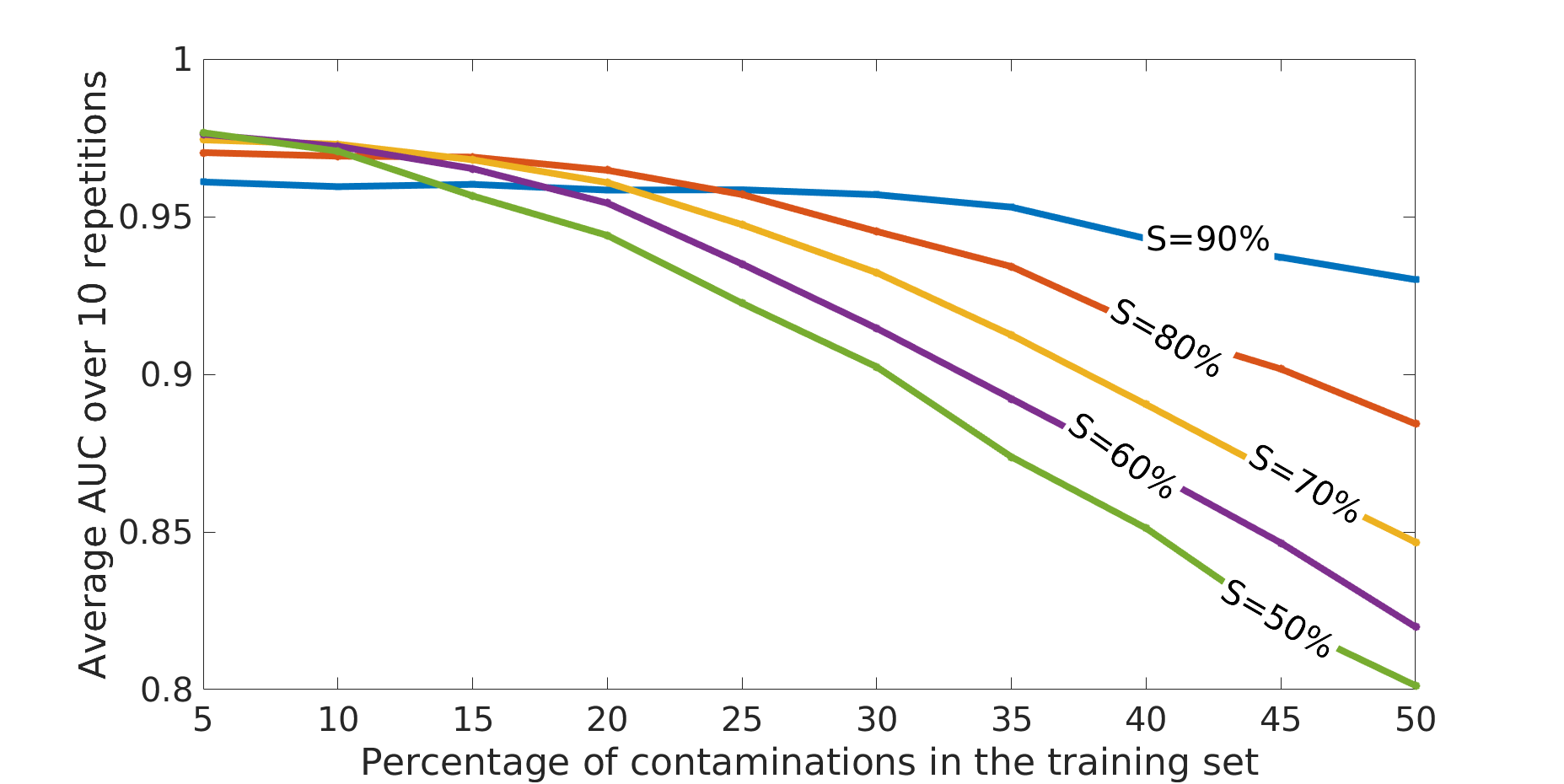}
\includegraphics[scale=.17]{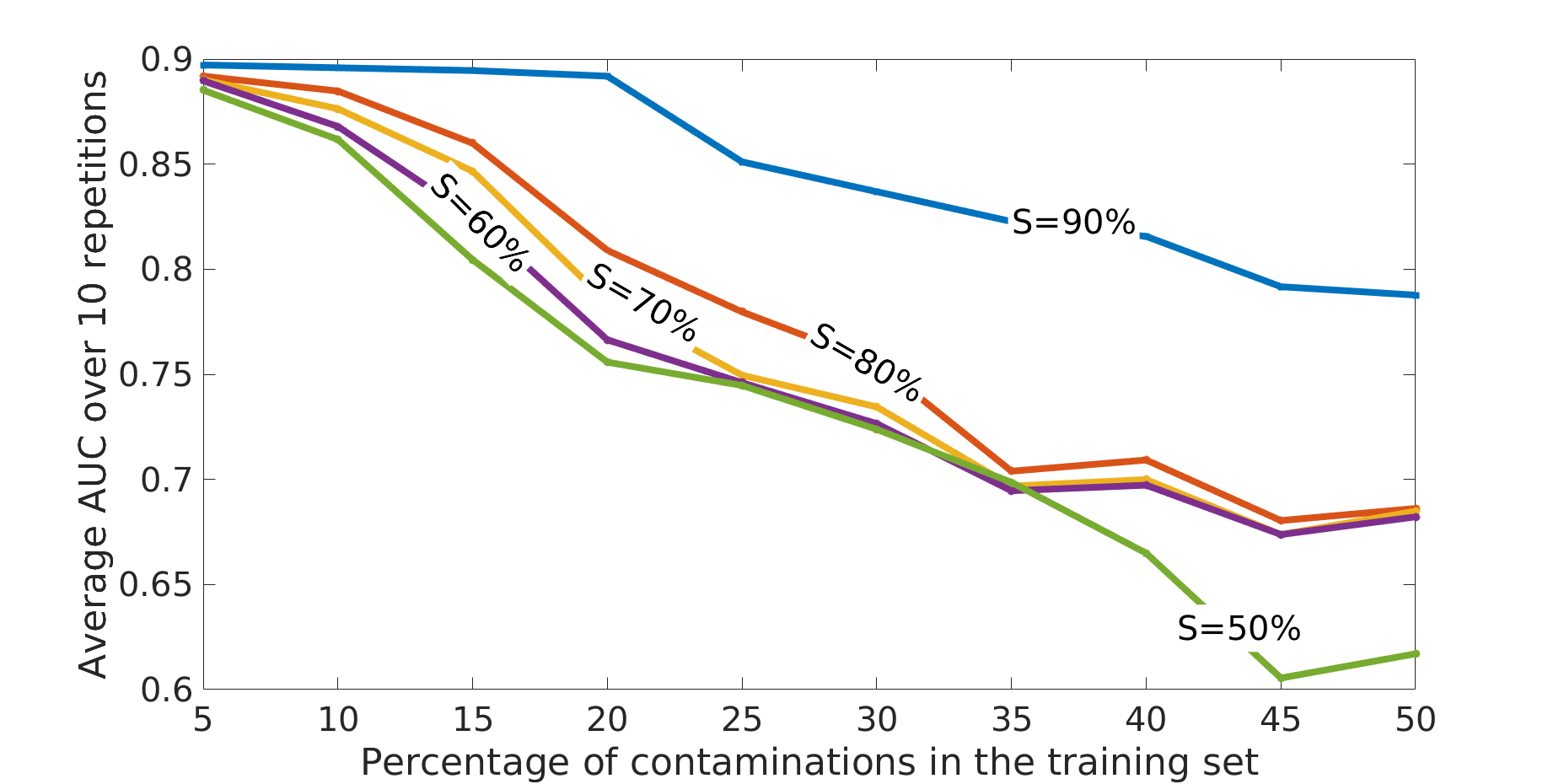}
\includegraphics[scale=.17]{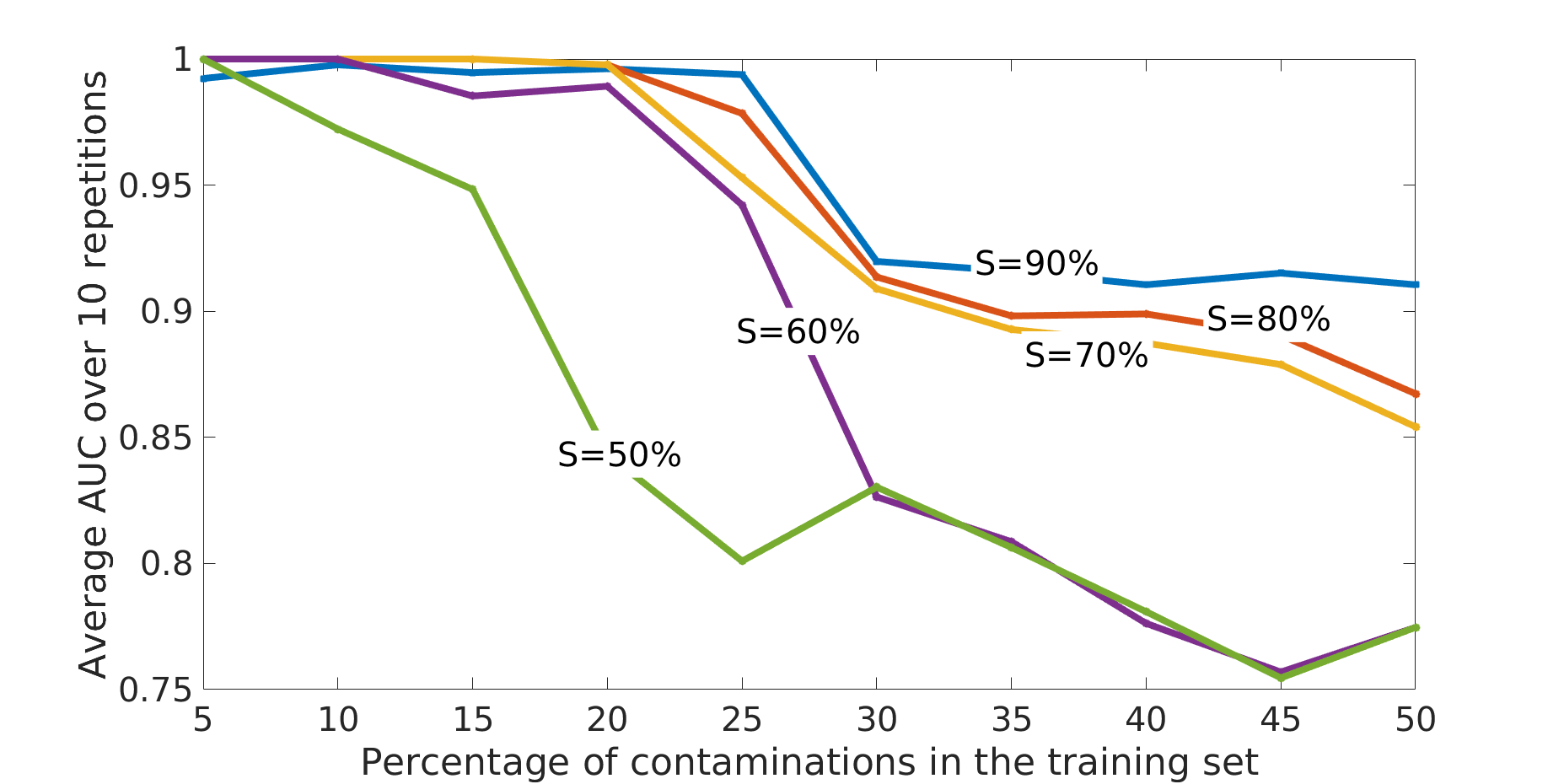}
\caption{The effect of varying sparseness on the performance of the proposed approach. From top to bottom: face data set, MNIST data set and the Coil-100 data set}
\label{var2}
\end{figure}

In this section, the effect of changing the regularisation parameters on the performance of the proposed methodology in presence of contaminations in the training set, varying from $5\%$ to $50\%$ in steps of $5\%$, is evaluated. Regarding the ridge regression formulation, the optimal Tikhonov regularisation parameter along with three other parameters and the case where no regularisation is imposed, is considered. For the sparse model, the degree of sparseness is varied from $50\%$ to $90\%$ in steps of $10\%$ where an $n\%$ sparsity corresponds to the case where $n\%$ of variables in $\boldsymbol\alpha$ are zero. The results corresponding to this experiment for the Tikhonov regularised formulation and the sparse model on the three data sets are presented in Fig. \ref{var1} and \ref{var2}, respectively. The plots correspond to random average AUC's over ten random splits of data into the training and test sets.

Regrading the Tikhonov regularised model, the worst performance corresponds to the case where no regularisation is applied. Increasing the regularisation parameter from zero towards the optimal value gradually enhances system robustness on all three data sets. Regarding the proposed sparse model, the sparser the solution towards a $90\%$ sparsity level, the better the robustness of the method is. This behaviour is confirmed on all three data sets where a maximum sparseness of $90\%$ achieves the best average overall performance over the entire range of contaminations. A common behaviour for both types of regularisation is that increasing the regularisation effect improves robustness against contaminations towards higher levels of corruption at the cost of a small decrease in performance for the lower percentages of contaminations. Nevertheless, the best average performance over the entire range of contaminations is obtained for the Tikhonov regularised model with the optimal regularisation parameter and the sparse model with a $90\%$ sparsity level, indicating the effectiveness of regularisation in improving the robustness of the kernel null-space approach.

\subsection{One-class classification in presence of contaminations in the training set}
In this experiment, the proposed regularised null-space kernel spectral regression methods are evaluated in a one-class classification problem where the model is trained on a training set of observations whose contamination is varied from $10\%$ to $50\%$ and then evaluated on a separate set of test samples. Such an evaluation scheme is commonly referred to as semi-supervised one-class classification \cite{Chandola:2009:ADS:1541880.1541882} where it is assumed that the training data incorporates labelled instances only for the normal class. In this experiment, the proposed methodology is compared against several other methods including:
\begin{itemize}
\item \textbf{Tikh} is the proposed robust spectral regression approach using a Tikhonov regularisation term.
\item \textbf{Spar} is the proposed robust spectral regression approach using a sparsity encouraging regularisation term.
\item \textbf{Org} corresponds to the one-class kernel null Foley-Sammon transform. The methods presented in \cite{DBLP:journals/corr/abs-1807-01085}, \cite{6619277} and \cite{8099922} theoretically obtain the same result and only differ in terms of computational complexity. As the proposed approach is based on a kernel null-space formulation, the methods in \cite{DBLP:journals/corr/abs-1807-01085,6619277,8099922} serve as a baseline to gauge improvements obtained using the proposed approach.
\item \textbf{SVDD} is the Support Vector Data Description approach to solve the one class classification problem \cite{Tax2004}. As a widely used method, it serves as a second baseline for comparison.
\item \textbf{GP} is derived based on the Gaussian process regression and approximate Gaussian process classification \cite{KEMMLER20133507} where in this work the predictive mean is used as the one class score.
\item \textbf{K-means} is the k-means clustering-based approach where k centres are assumed for the target observation. The novelty score of a sample is defined as the minimum distance of a query to cluster centres. The optimum value for parameter $k$ is experimentally set to be 5 to achieve the best average performance over all data sets.
\item \textbf{FB} corresponds to the feature bagging algorithm \cite{DBLP:conf/kdd/LazarevicK05} which detects outliers by bagging anomaly scores where the scores are generated by different individual outlier detection models operating on random subsets of input features. For each outlier detection model, the abnormal score is derived using a small subset from feature set. The base outlier detection method is that of the Local Outlier Factor \cite{Breunig00lof:identifying}. 
\item \textbf{Parzen} corresponds to the non-parametric density estimation approach based on Parzen-window estimators with Gaussian kernels \cite{1047476} and works by sliding a window centred at each sample, and computing a probability density function for each label. Samples for which the maximum probability is less than a threshold are identified as novelties. 
\end{itemize}

\subsubsection{Implementation details}
For the Org method, the implementation corresponding to the work in \cite{DBLP:journals/corr/abs-1807-01085} is employed as it is the most efficient among other variants of the kernel null-space technique. The SVDD method is based on the implementation provided in the data description toolbox \cite{Ddtools2018}. The GP method is based on the implementation available publicly \footnote{https://github.com/erodner/gpocc}. Regarding the FB and Parzen approaches the implementations provided in a supplement website \footnote{https://github.com/gokererdogan/OutlierDetectionToolbox.} are utilised. The K-means clustering approach is based on its implementation in Matlab 2017b. No pre-processing is applied on the features other than normalising them to have a unit $l_2$-norm.

In the experiments on each data set, the data is randomly divided into the train and test sets and the percentage of contaminations is increased from $10\%$ to $50\%$ of the total training data composed of positive and negative samples. Note that a $50\%$ contamination in the training set represents a quite high degree of data corruption. The performances are then reported as the average AUC's over 10 repetitions of random splitting of the data into the train and test sets.
\begin{figure*}
\centering
\includegraphics[scale=.18]{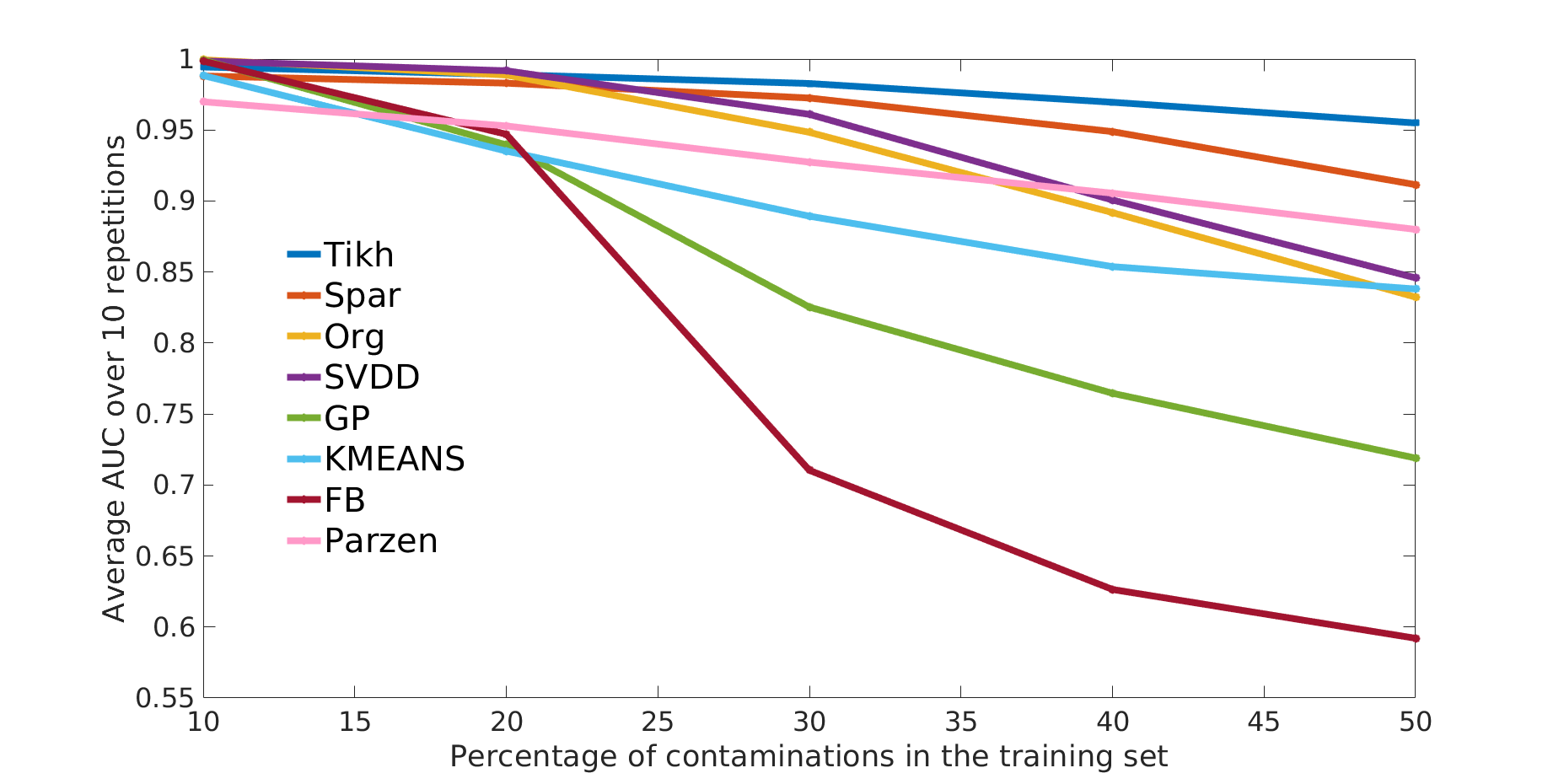}
\includegraphics[scale=.18]{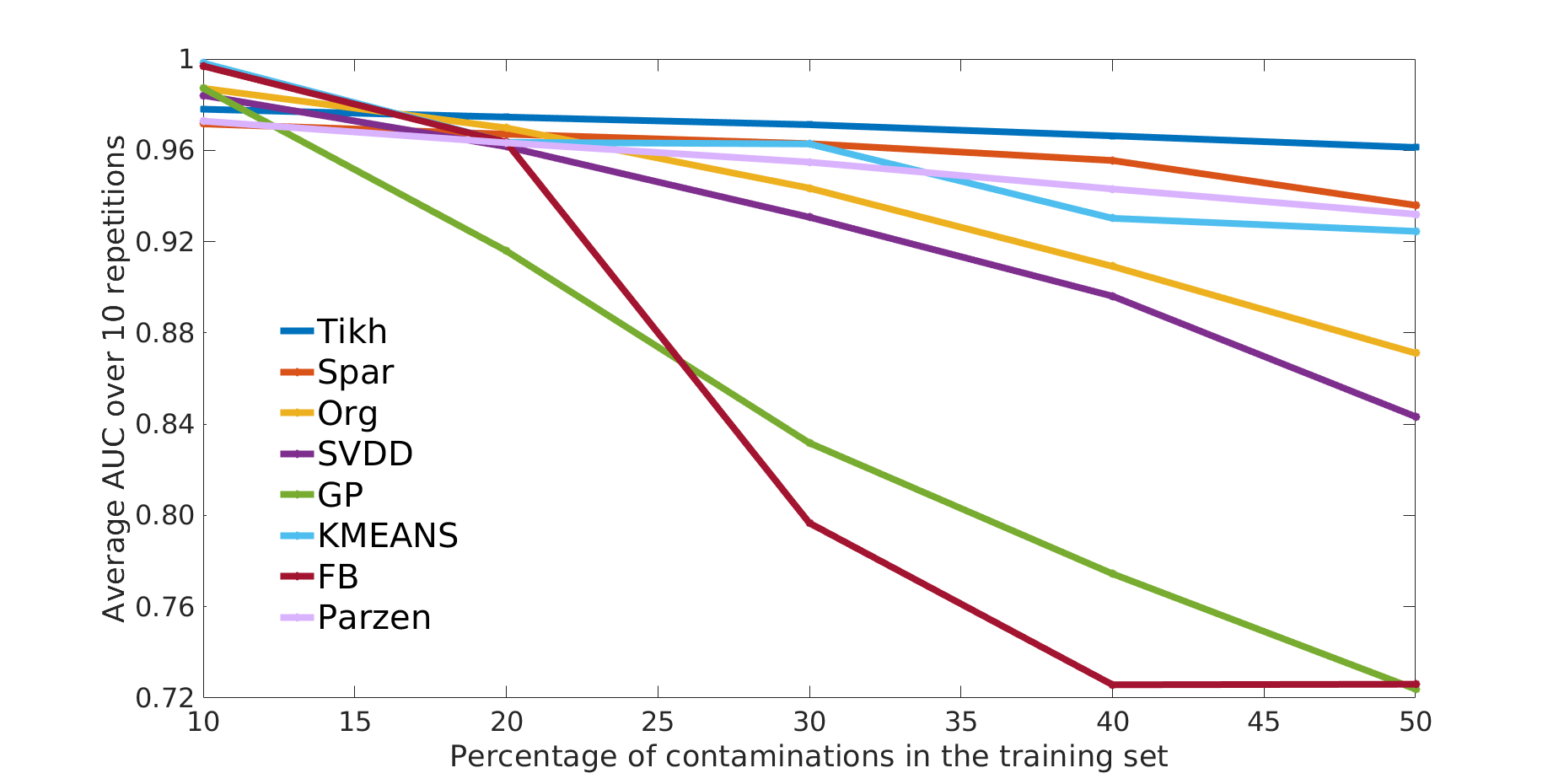}
\includegraphics[scale=.18]{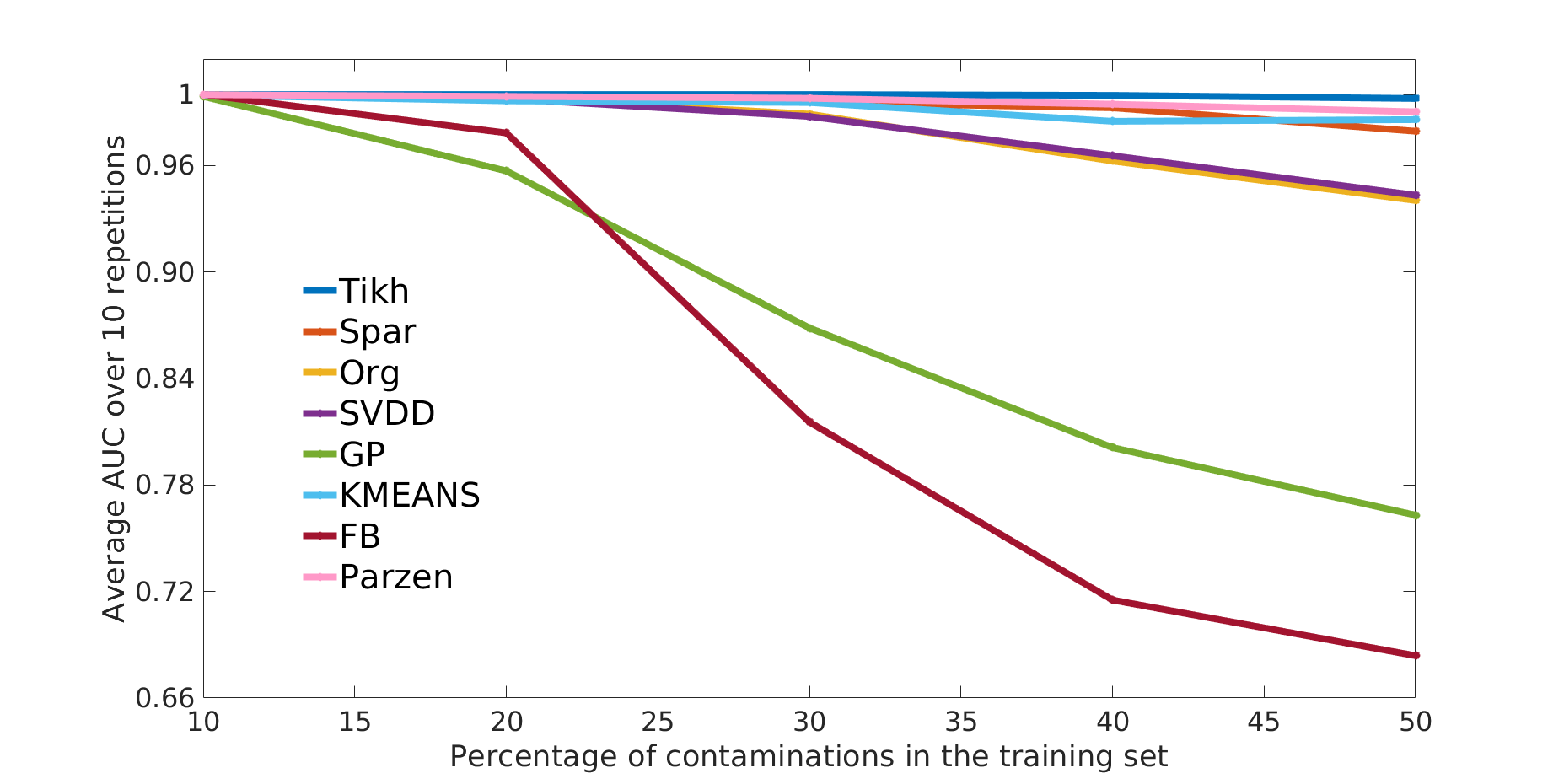}
\includegraphics[scale=.18]{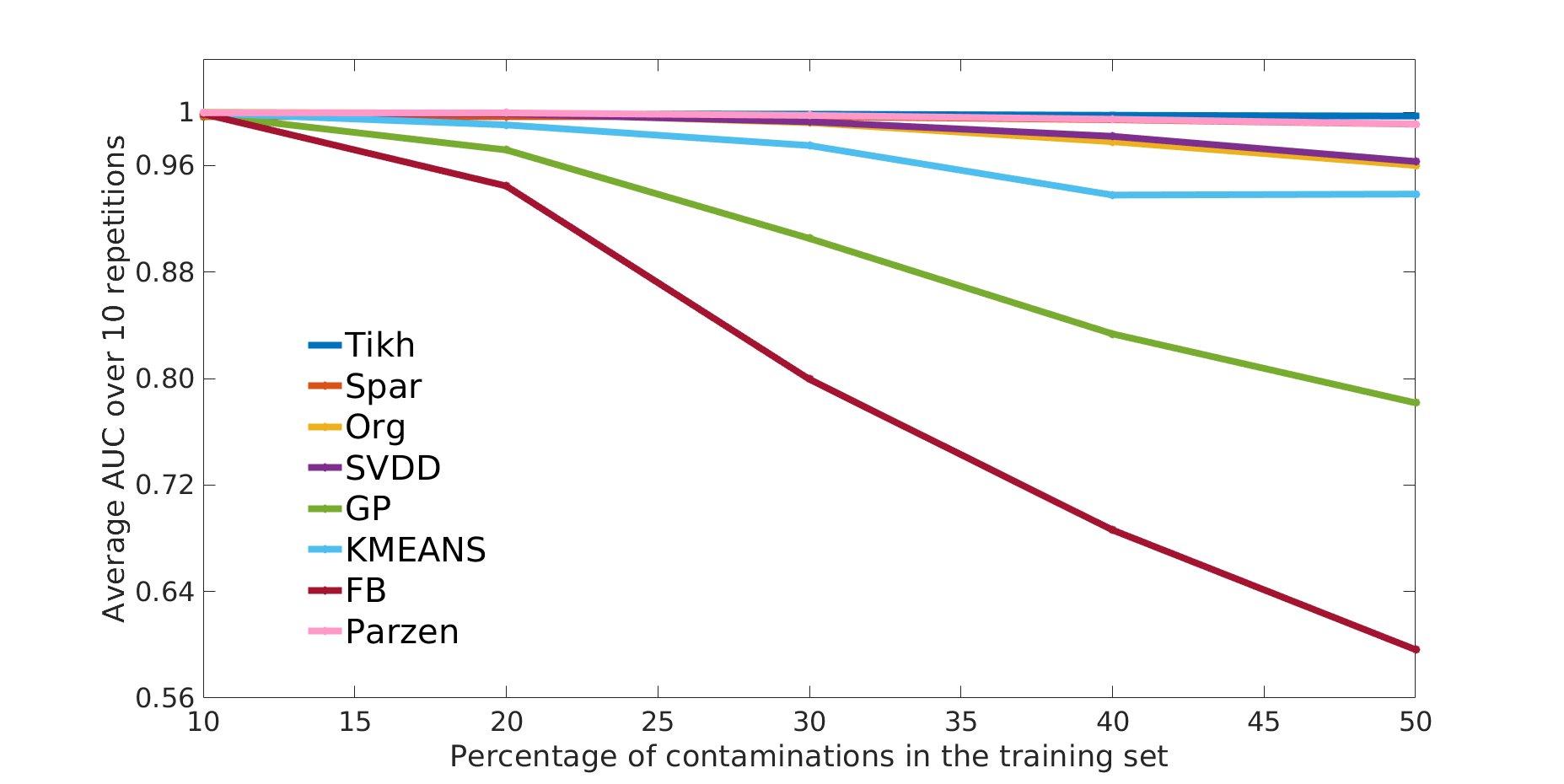}
\includegraphics[scale=.18]{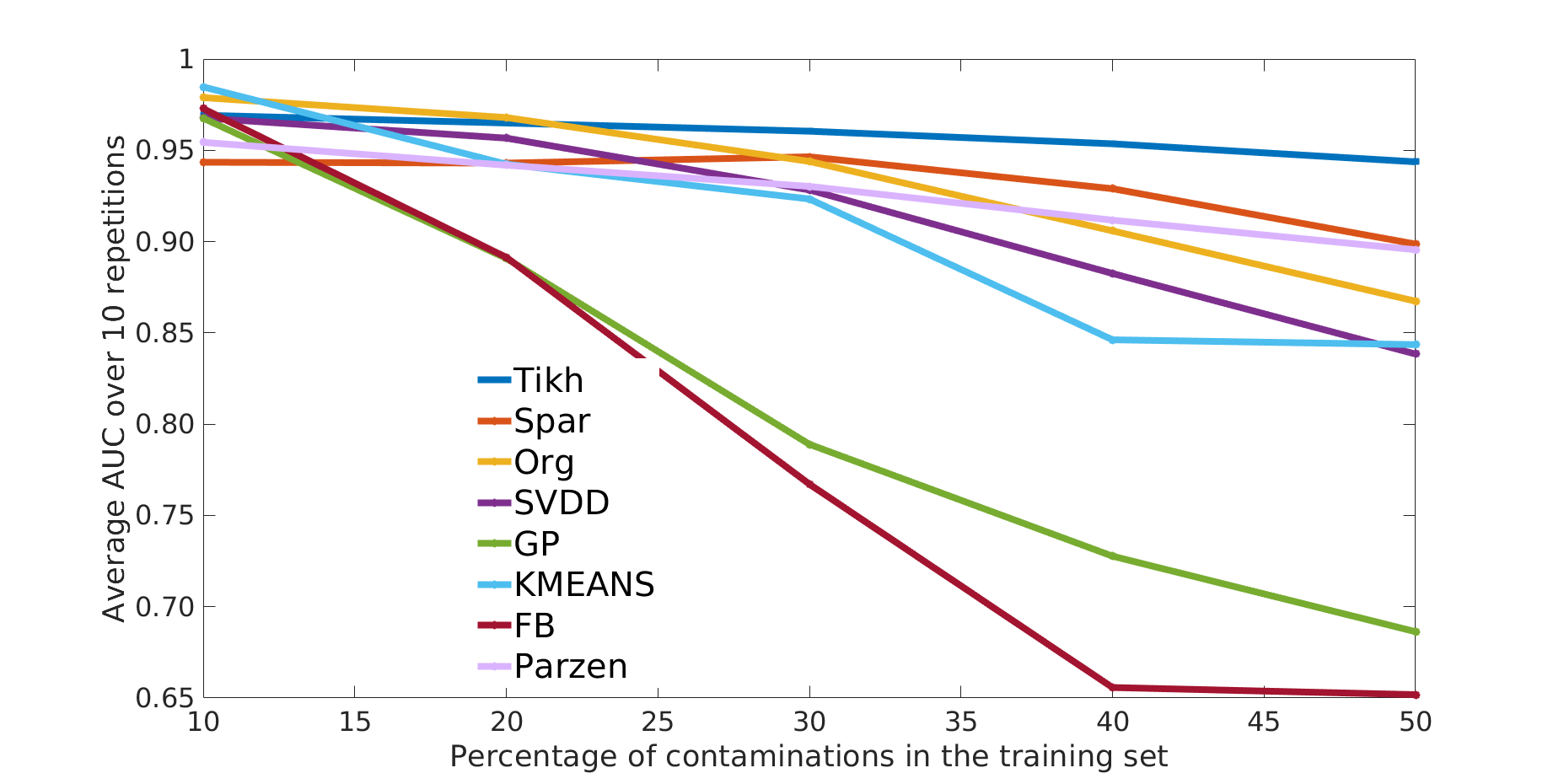}
\includegraphics[scale=.18]{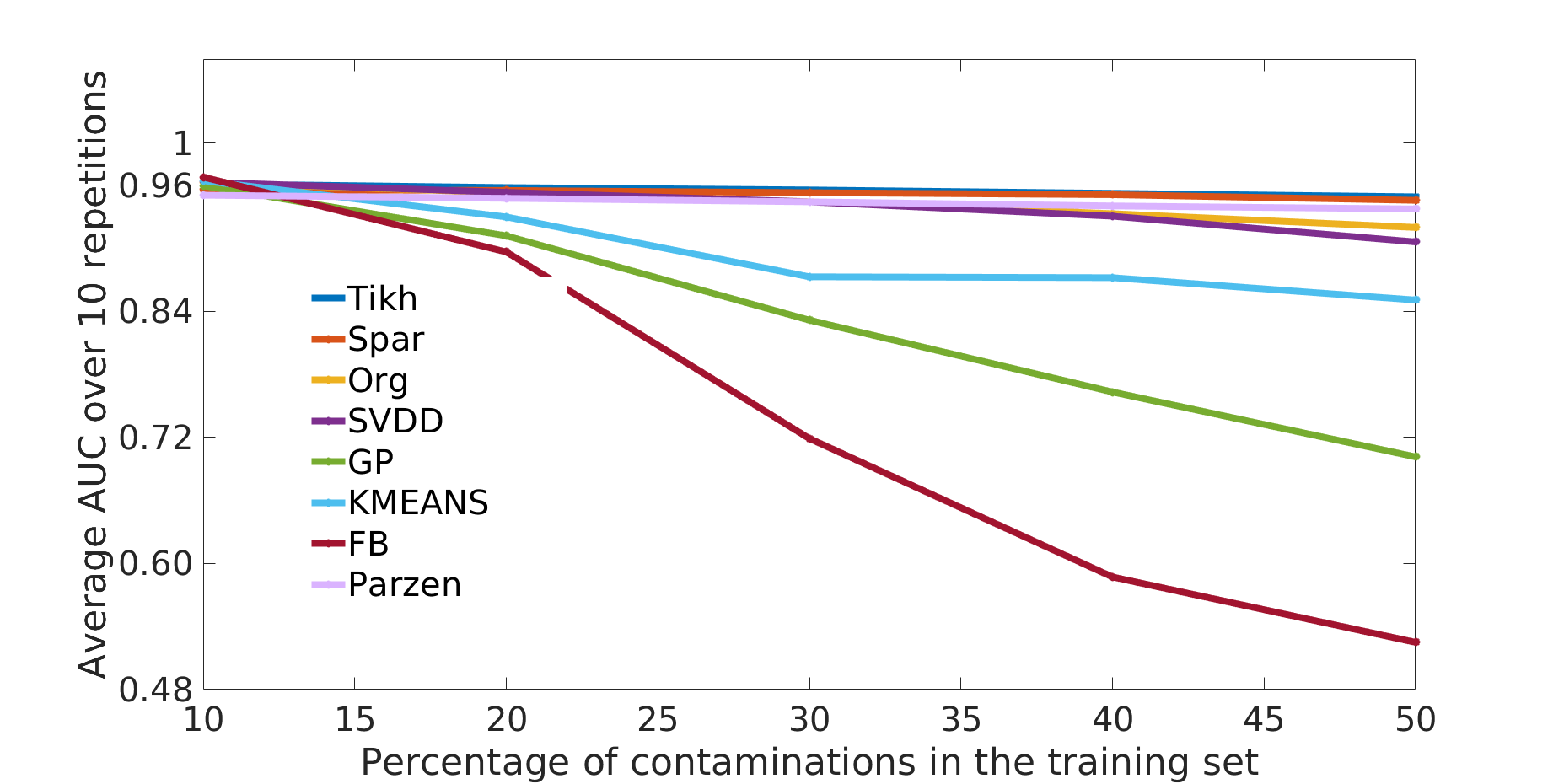}
\includegraphics[scale=.18]{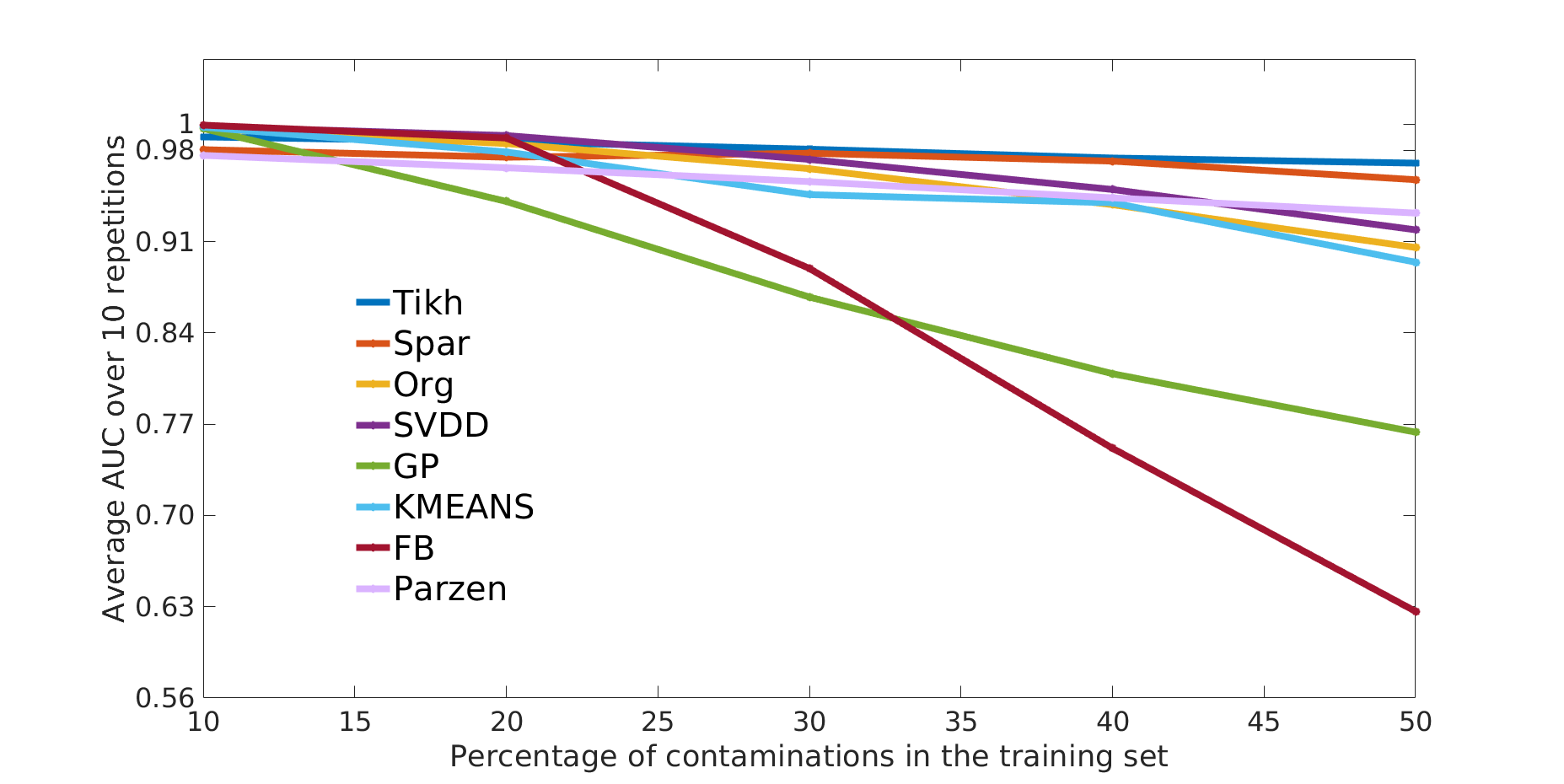}
\includegraphics[scale=.18]{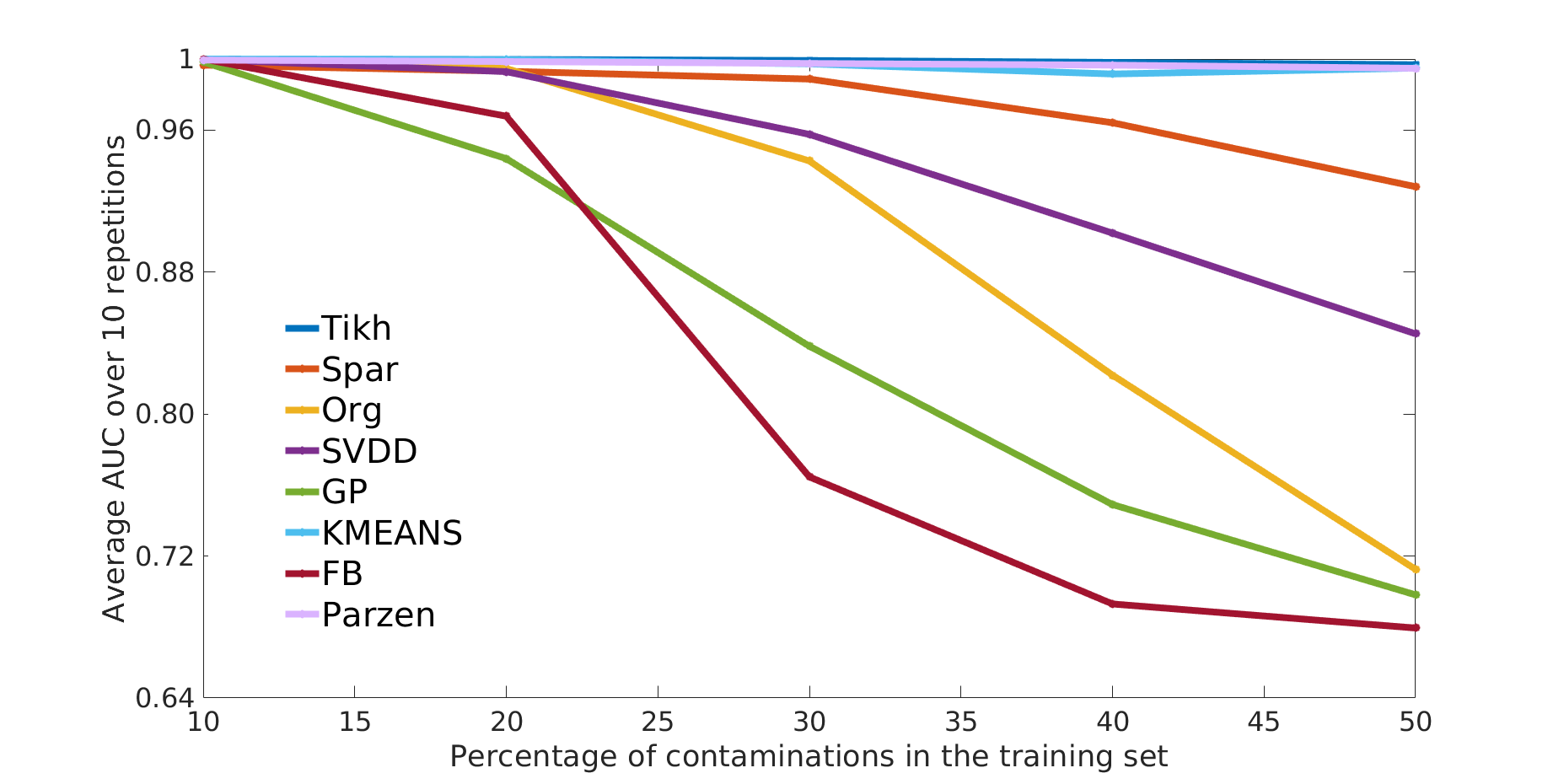}
\includegraphics[scale=.18]{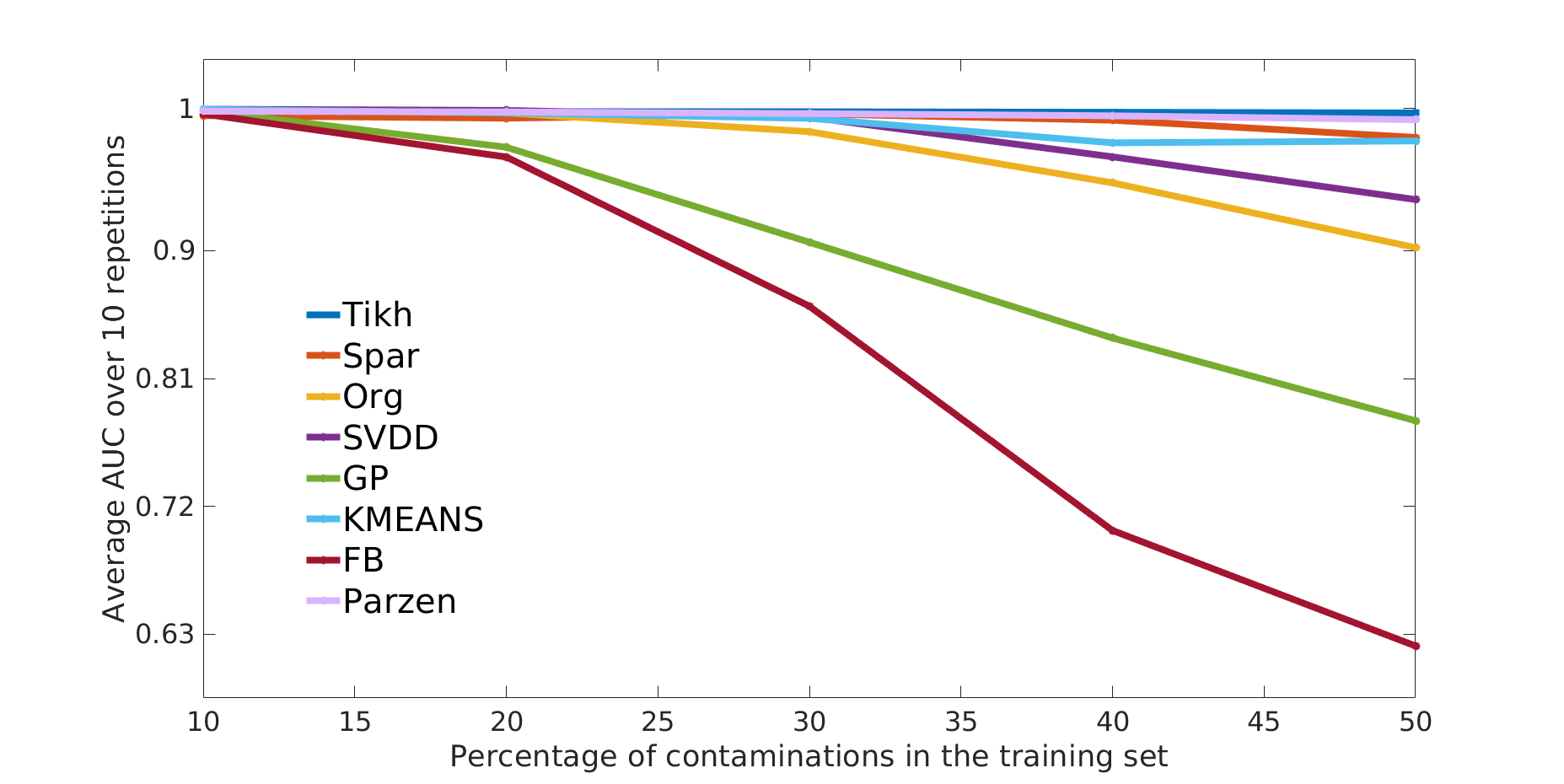}
\includegraphics[scale=.18]{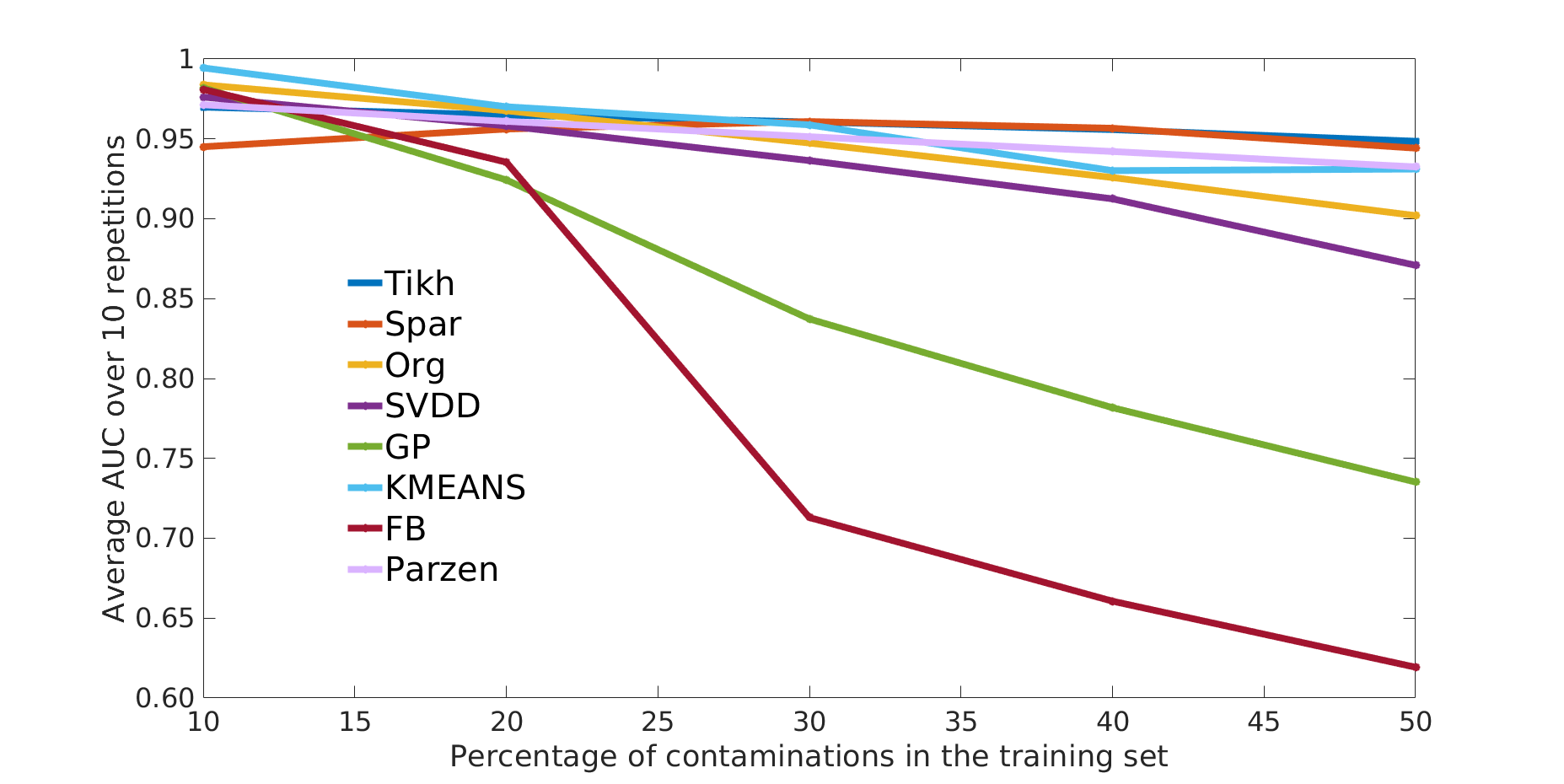}
\caption{Average performance in terms of AUC's for different methods on the face data set for 10 different subjects.}
\label{fres1}
\end{figure*}

\begin{figure}
\centering
\includegraphics[scale=.18]{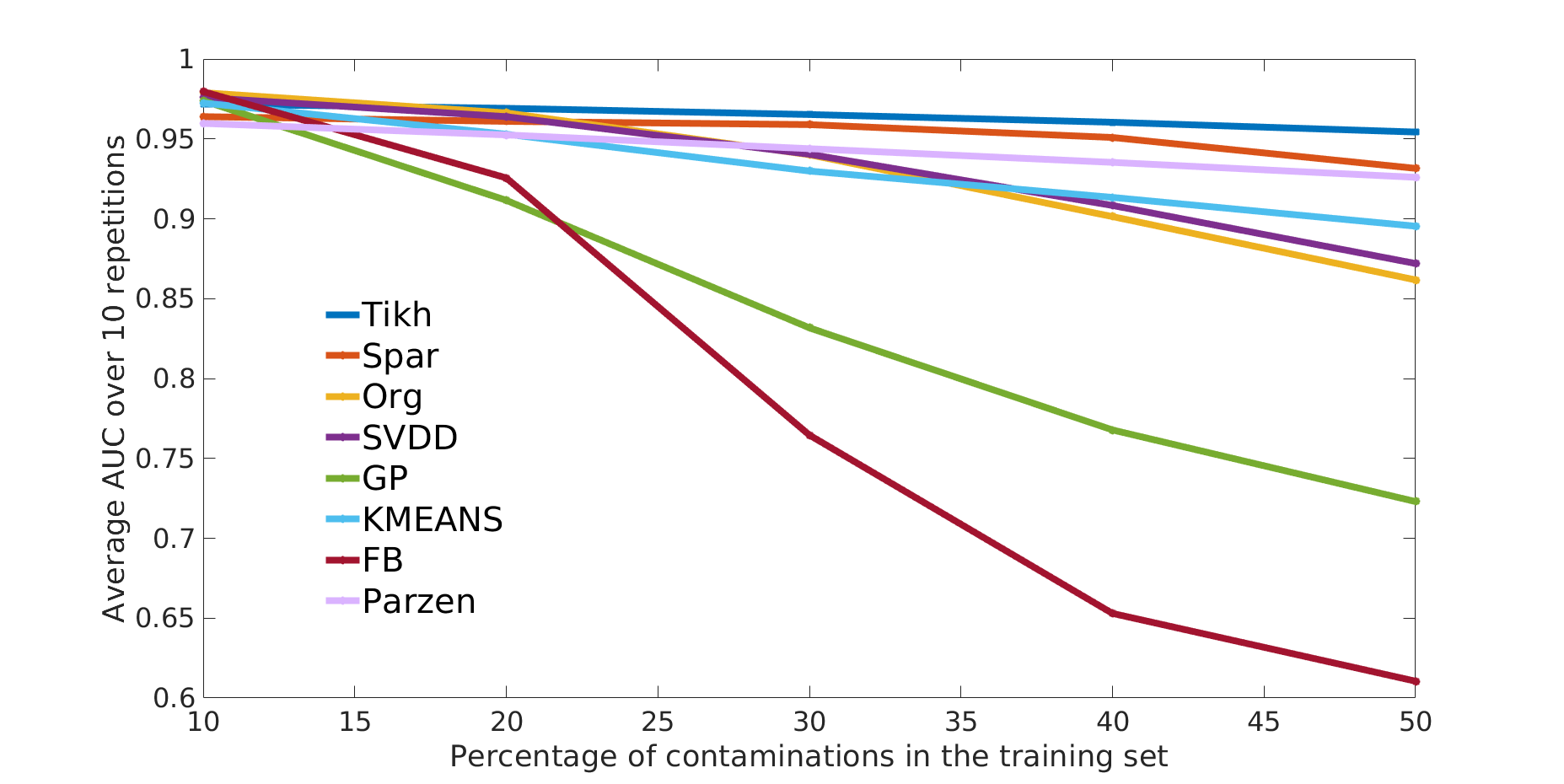}
\caption{Average AUC's for different methods over all subjects on the face data set.}
\label{fres2}
\end{figure}

\begin{figure}
\centering
\includegraphics[scale=.18]{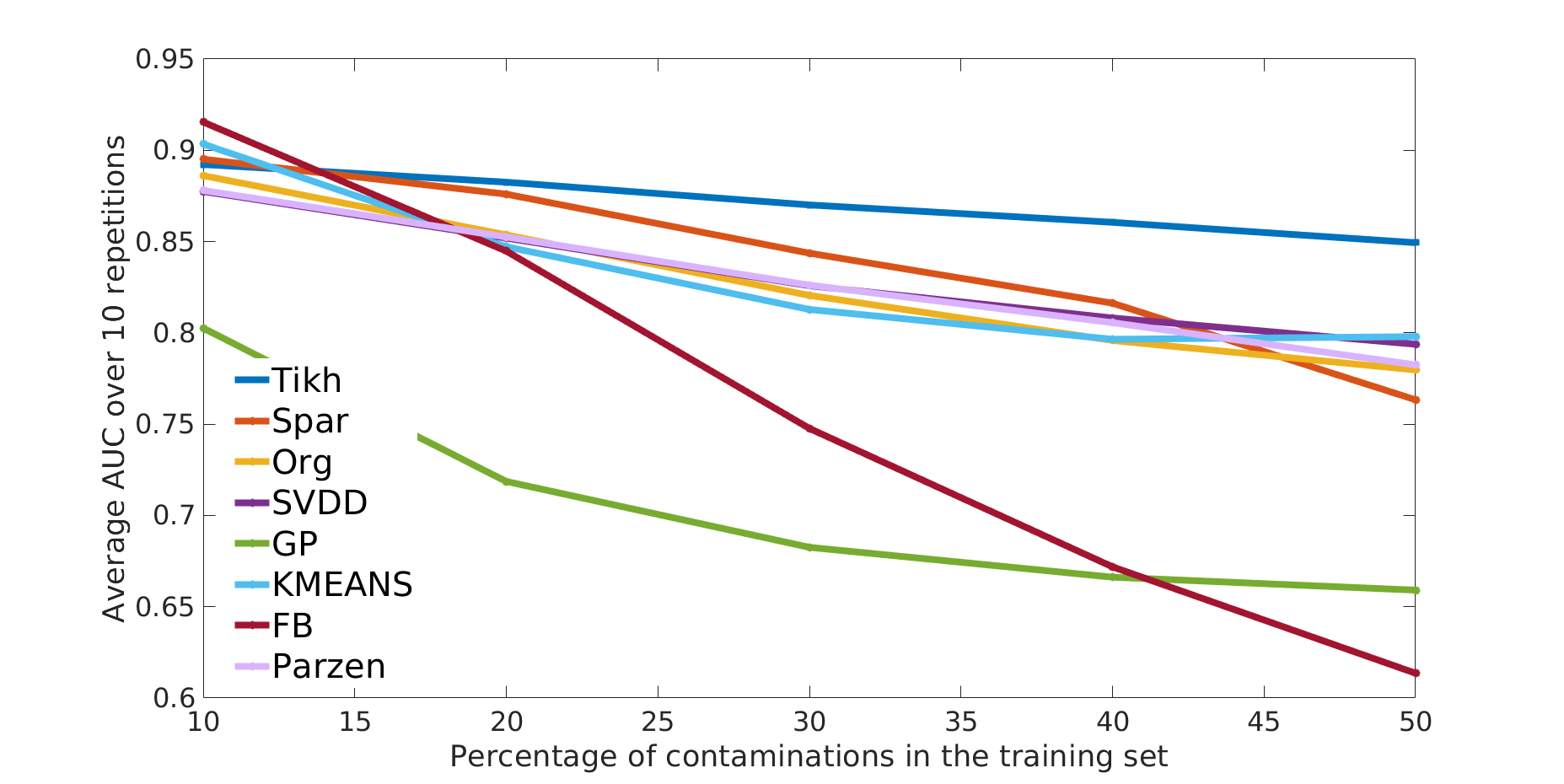}
\caption{Average AUC's for different methods on the MNIST data set.}
\label{mnistres}
\end{figure}

\begin{figure}
\centering
\includegraphics[scale=.18]{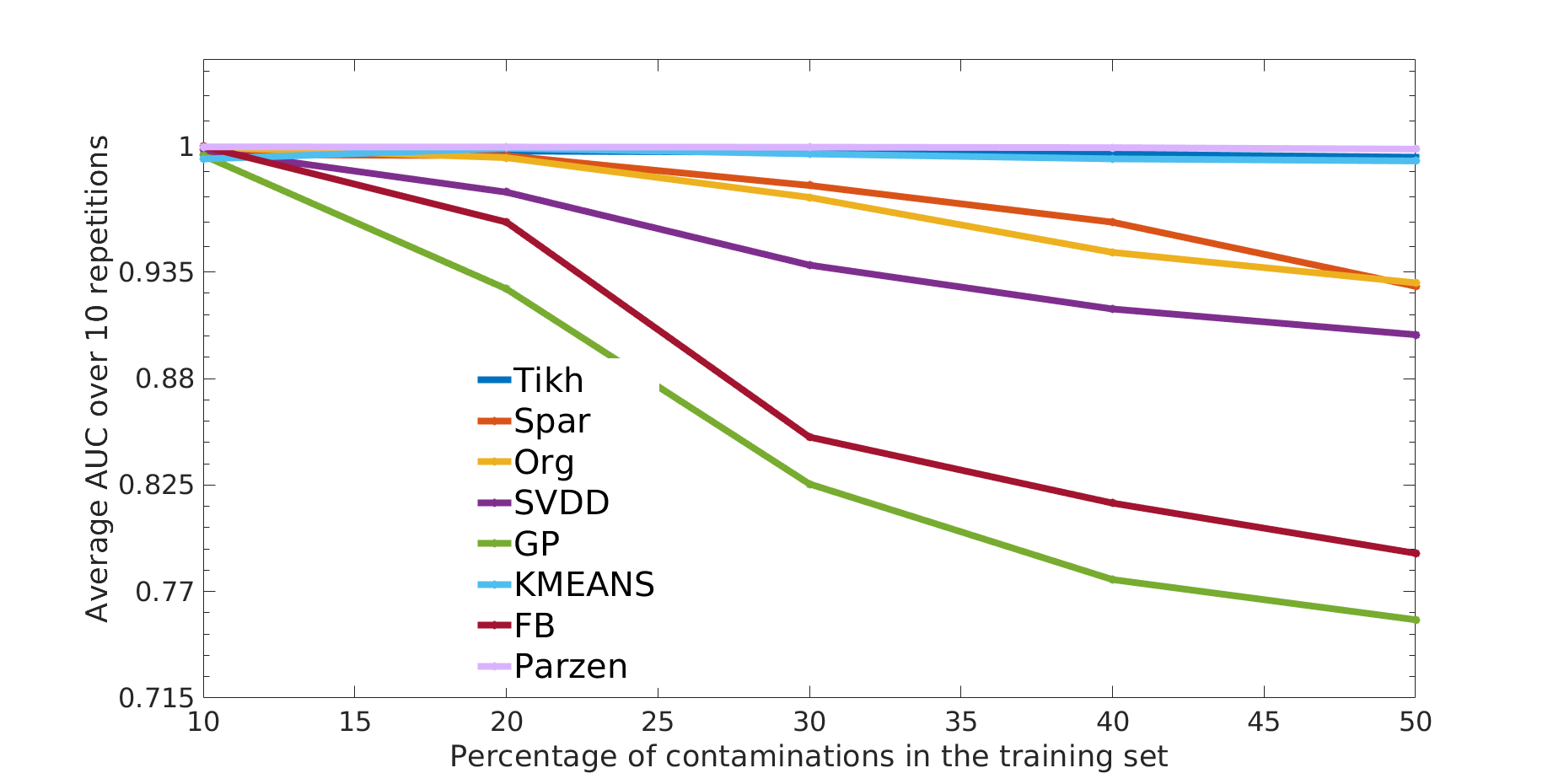}
\caption{Average AUC's for different methods on on the Coil-100 data set.}
\label{coilres}
\end{figure}

\subsubsection{Results on the face data set}
The results in terms of average AUC's over ten random splitting of the data into the train and test sets on the face data set are plotted in Fig. \ref{fres1} for each subject separately. The overall performance over all subjects is also provided in Fig. \ref{fres2}. The average AUC's (in percentage) for different subjects over the whole range of contaminations (from $10\%$ to $50\%$) are also reported in Table \ref{frestab}.

From the figures and the table, it can be observed that the best performing method, on average, over the full range of contaminations is the proposed approach based on Tikhonov regularisation. Moreover, in terms of the average performance over all subjects, the second best performing method is that of the proposed sparse approach. The improvement obtained using the proposed approach over the baseline method \cite{DBLP:journals/corr/abs-1807-01085,6619277,8099922} is more than $3\%$ on average using a Tikhonov regularisation and over $2\%$ on average using a sparse regularisation. However, as the percentage of contaminations in the training set increases, the effectiveness of the proposed regularisation-based approach becomes more evident reaching more than $10\%$ of improvement in terms of AUC for a $50\%$ contamination in the training set.

\begin{table*}
\footnotesize
\renewcommand{\arraystretch}{1.2}
\caption{Average performance (in terms of AUC ($\%$)) of different methods in a one-class classification scenario on the face data set over the full range of contamination percentages ($10\%-50\%$)}
\label{Characteristics}
\centering
\begin{tabular}{lcccccccc}
\hline
\textbf{Method} & \textbf{Tikh} & \textbf{Spar} & \textbf{Org} & \textbf{SVDD} & \textbf{GP} & \textbf{K-means} & \textbf{FB} & \textbf{Parzen} \\
\hline
Subject $\boldsymbol{\#}1$ &96.20 &94.71 &90.09 &90.74 &80.23 &87.99 &78.58 &90.01\\
Subject $\boldsymbol{\#}2$ &97.03 &95.86 &93.61 &92.31 &84.66 &95.58 &84.18 &95.32\\
Subject $\boldsymbol{\#}3$ &99.95 &99.30 &97.80 &97.88 &87.77 &99.25 &83.86 &99.63\\
Subject $\boldsymbol{\#}4$ &99.82 &99.51 &98.59 &98.72 &89.80 &96.83 &80.51 &99.66\\
Subject $\boldsymbol{\#}5$ &95.84 &93.21 &93.28 &91.47 &81.23 &90.80 &78.78 &92.67\\
Subject $\boldsymbol{\#}6$ &95.51 &95.20 &94.24 &93.93 &83.35 &89.79 &73.90 &94.39\\
Subject $\boldsymbol{\#}7$ &98.04 &97.24 &95.83 &96.63 &87.55 &95.13 &85.12 &95.48\\
Subject $\boldsymbol{\#}8$ &99.88 &97.41 &89.43 &93.95 &84.55 &99.67 &82.10 &99.74\\
Subject $\boldsymbol{\#}9$ &99.77 &99.13 &96.60 &97.89 &89.92 &98.86 &82.96 &99.60\\
Subject $\boldsymbol{\#}10$ &95.68 &94.68 &94.55 &92.86 &85.76 &94.80 &77.82 &95.00\\
All Subjects &96.80 &95.89 &93.61 &93.65 &84.80 &94.30 &78.50 &94.82\\
\hline
\label{frestab}
\end{tabular}
\end{table*}

\subsubsection{Results on the MNIST data set}
The results corresponding to different methods on the MNIST data set are provided in Fig. \ref{mnistres}. From the figure it can be observed that the most robust method against contaminations in the training set is the proposed approach based on Tikhonov regularisation. The superiority of this formulation over other alternatives becomes more evident as more contaminations are included in the training set. The second best performing method, on average, similar to the experiments on the face data set is again the proposed approach using a sparse regularisation term.

\subsubsection{Results on the Coil-100 data set}
The results corresponding to different methods on the Coil-100 data set are presented in Fig. \ref{coilres} in terms of average AUC's. From the figure it can be observed that, interestingly, the two best performing methods are those of Parzen window and the proposed approach based on Tikhonov regularisation. Nevertheless, the proposed sparse model is not performing much inferior with respect to the top performers on this data set. The results corresponding to the MNIST and Coil-100 data set are also summarised in Table \ref{coilmnistres}. As can be observed from the table, the two best performing methods on the MNIST data sets are the proposed methods based on Tikhonov and Sparse regularisation. In particular, the proposed approach based on Tikhonov regularisation, on average, is nearly $5\%$ better than the original kernel null-space method \cite{DBLP:journals/corr/abs-1807-01085,6619277,8099922}. On the Coil-100 data set, interestingly, the Parzen window method achieves the best average performance followed by the Tikhonov regularisation-based approach.
\begin{table*}
\footnotesize
\renewcommand{\arraystretch}{1.2}
\caption{Summary of the average performance (in terms of AUC ($\%$)) of different methods in a one-class classification scenario on the MNIST and Coil-100 data sets over the full range of contamination percentages ($10\%-50\%$)}
\label{Characteristics}
\centering
\begin{tabular}{lcccccccc}
\hline
\textbf{Method} & \textbf{Tikh} & \textbf{Spar} & \textbf{Org} & \textbf{SVDD} & \textbf{GP} & \textbf{K-means} & \textbf{FB} & \textbf{Parzen} \\
\hline
MNIST &87.10 &83.89 &82.72 &83.14 &70.58 &83.16 &75.87 &82.90\\
Coil-100 &99.64 &97.18 &96.83 &94.63 &85.57 &99.48 &88.32 &99.93\\
\hline
\end{tabular}
\label{coilmnistres}
\end{table*}

\subsection{One-class classification in presence of contaminations in the training set-known fraction of outliers}
\begin{figure}[t]
\centering
\includegraphics[scale=.17]{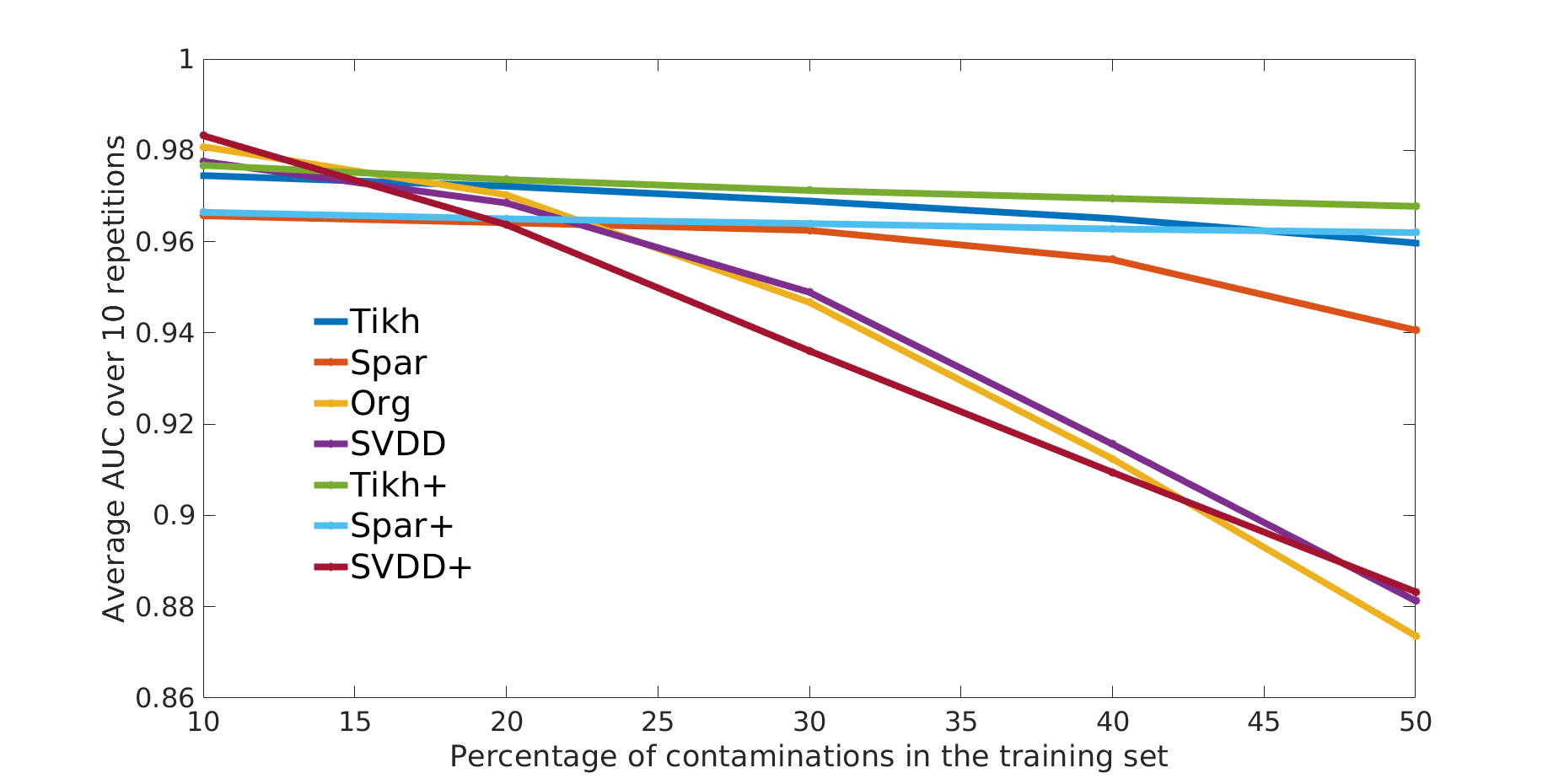}
\caption{Average AUC on the face data set over all subjects-known fraction of contaminations in the training set.}
\label{faceknown}
\end{figure}

\begin{figure}[t]
\centering
\includegraphics[scale=.17]{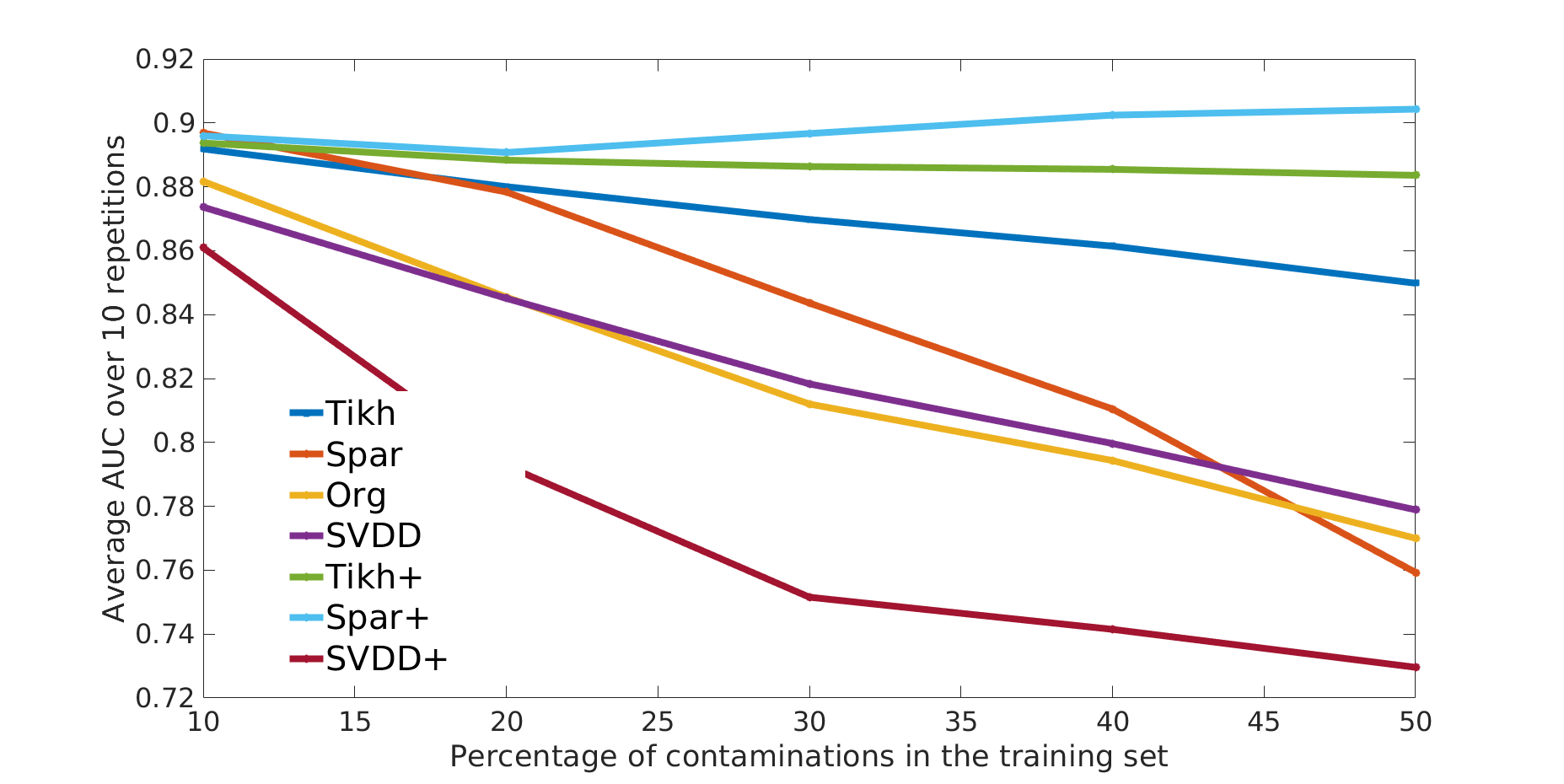}
\caption{Average AUC on the MNIST data set-known fraction of contaminations in the training set.}
\label{mnistknown}
\end{figure}

\begin{figure}[t]
\centering
\includegraphics[scale=.17]{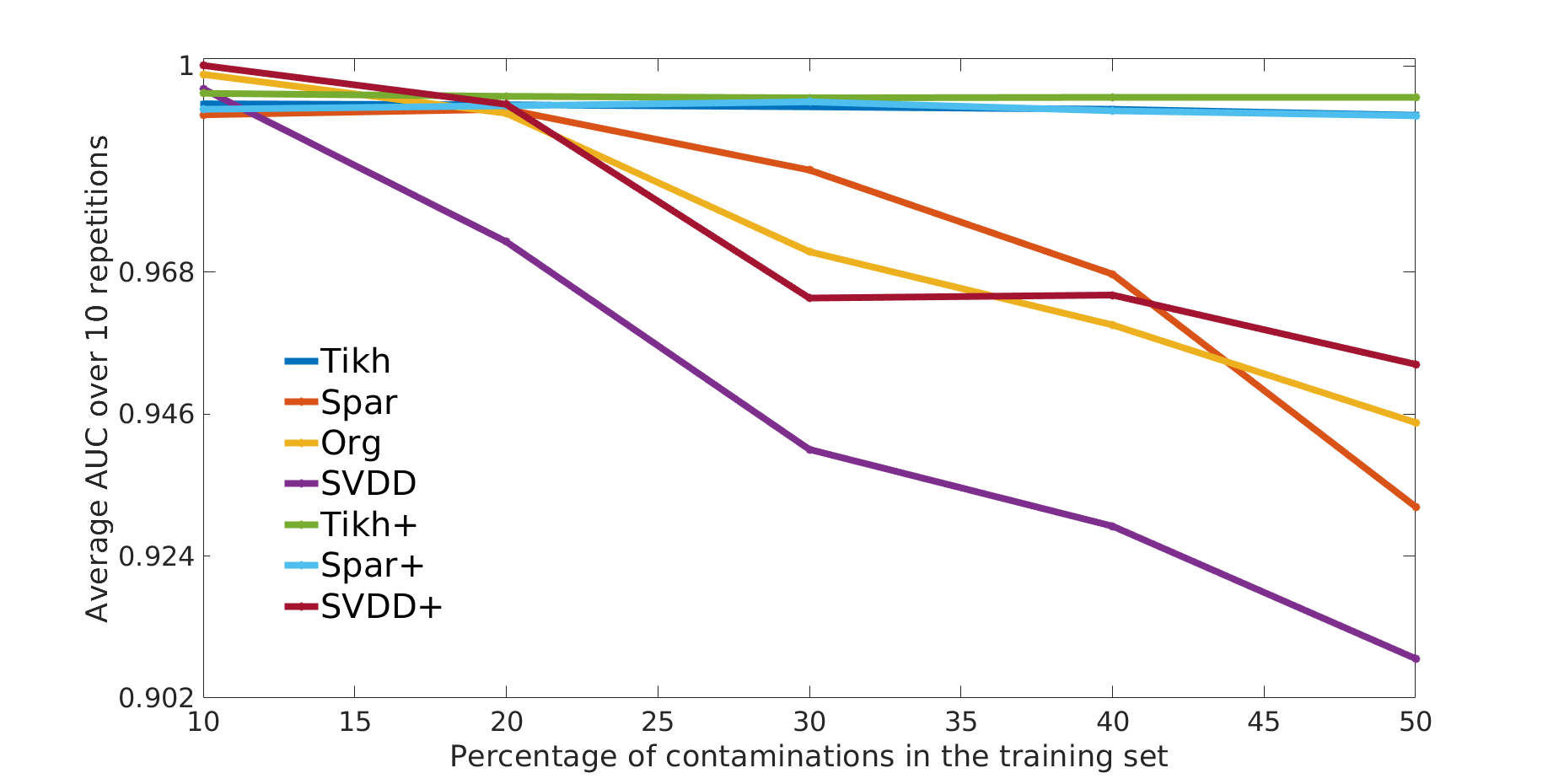}
\caption{Average AUC on the Coil-100 data set-known fraction of contaminations in the training set.}
\label{coilknown}
\end{figure}

In the previous experiments, it was assumed that the fraction of contaminations in the training set is not known. In case such information is available, it can be utilised in the context of the proposed approaches. This is realised by detecting contaminations in the training set and forming a second class corresponding to counter-examples which is used to refine decision boundaries. In this experiment, the number of outliers in the training set is provided as an additional input parameter and the results are compared to the  proposed naive methods where no such information is available. Among other techniques, only the SVDD method is able to use such information for one-class learning. As a result, the methods included in this experiment are:
\begin{itemize}
\item \textbf{Tikh} is the proposed robust spectral regression approach using a Tikhonov regularisation term without making use of the information regarding the fraction of outliers in the training set.
\item \textbf{Spar} is the proposed robust spectral regression approach using a sparsity encouraging regularisation term without making use of the information regarding the fraction of outliers in the training set.
\item \textbf{Tikh+} is the proposed robust spectral regression approach based on Tikhonov regularisation provided by the information regarding the fraction of outliers in the training set.
\item \textbf{Spar+} is the proposed robust sparse spectral regression approach using the information regarding the fraction of outliers in the training set.
\item \textbf{Org} corresponds to the one-class kernel null Foley-Sammon transform \cite{DBLP:journals/corr/abs-1807-01085,6619277,8099922}.
\item \textbf{SVDD} is the Support Vector Data Description approach to solve the one class classification problem without making use of the information regarding the fraction of outliers in the training set.
\item \textbf{SVDD+} is the Support Vector Data Description approach making use of the information regarding the fraction of outliers in the training set.
\end{itemize}
The results corresponding to this experiment are presented in Figures \ref{faceknown}, \ref{mnistknown} and \ref{coilknown} for the face, MNIST and Coil-100 data sets and also summarised in Table \ref{facecoilmnistknown}.

\begin{table*}
\footnotesize
\renewcommand{\arraystretch}{1.2}
\caption{Summary of the average performance (in terms of AUC ($\%$)) of different methods in a one-class classification scenario on the face, MNIST and Coil-100 data sets over the full range of contamination percentages ($10\%-50\%$) when the fraction of contaminations is known.}
\label{Characteristics}
\centering
\begin{tabular}{lccccccc}
\hline
\textbf{Method} & \textbf{Tikh} & \textbf{Spar} & \textbf{Tikh+} & \textbf{Spar+} & \textbf{Org} & \textbf{SVDD} & \textbf{SVDD+} \\
\hline
face & 96.80 & 95.78 & 97.17 & 96.40 & 93.67 & 93.84 &93.51 \\
MNIST & 87.06 & 83.77 & 88.75 & 89.80 & 82.07 & 82.31 &77.53 \\
Coil-100 & 99.34 & 97.37 & 99.52 & 99.33 & 97.34 & 94.93 & 97.52 \\
\hline
\end{tabular}
\label{facecoilmnistknown}
\end{table*}
From the figures and the table, a number of conclusions can be drawn. First, it can be observed that the proposed methods can effectively utilised the information regarding the number of contaminations in the training set on all three data sets. This can be verified by the fact that Tikh+ and Spar+ perform better that their naive versions Tikh and Spar on all the three data sets examined. Second, the improvement obtained in Tikh+ and Spar+ is more pronounced with an increase in number of non-target observations in the training set which is indicative of the fact that in the proposed methodology non-target samples can be effectively detected. Third, it can also be observed that, interestingly, on the MNIST data set the additional information in terms of number of negative samples in the training set may even result in a boost in the performance of Spar+ method with an increase in the number of contaminations. This is reflected in the average AUC of Spar+ method corresponding to a $50\%$ corruption being higher than that of a $10\%$ corruption in the data set. 
\subsection{Unsupervised observation ranking}
\begin{figure}[t]
\centering
\includegraphics[scale=.17]{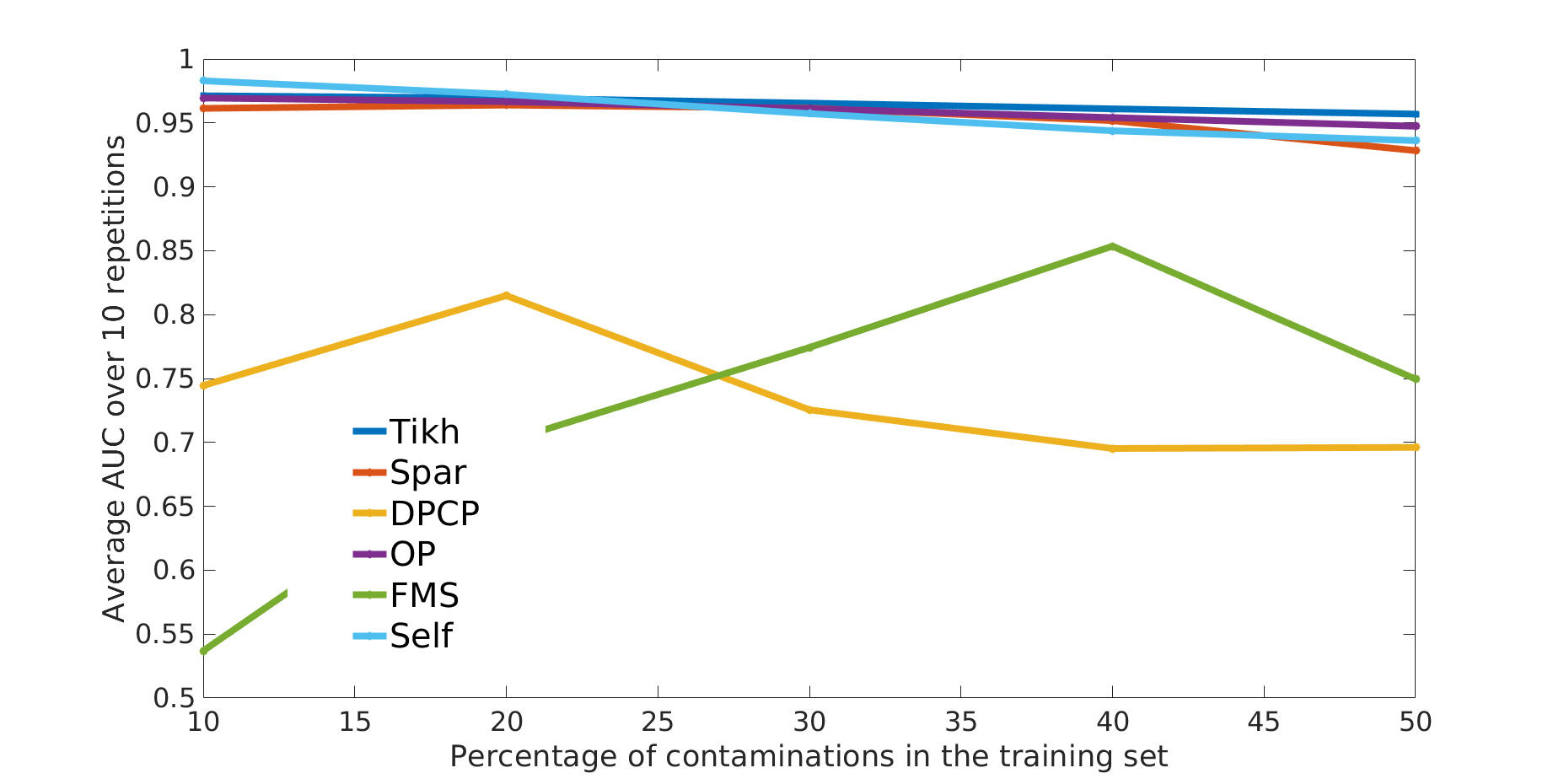}
\caption{Average AUC over all subjects on the face data set corresponding to a ranking of training samples.}
\label{unface}
\end{figure}

\begin{figure}[t]
\centering
\includegraphics[scale=.17]{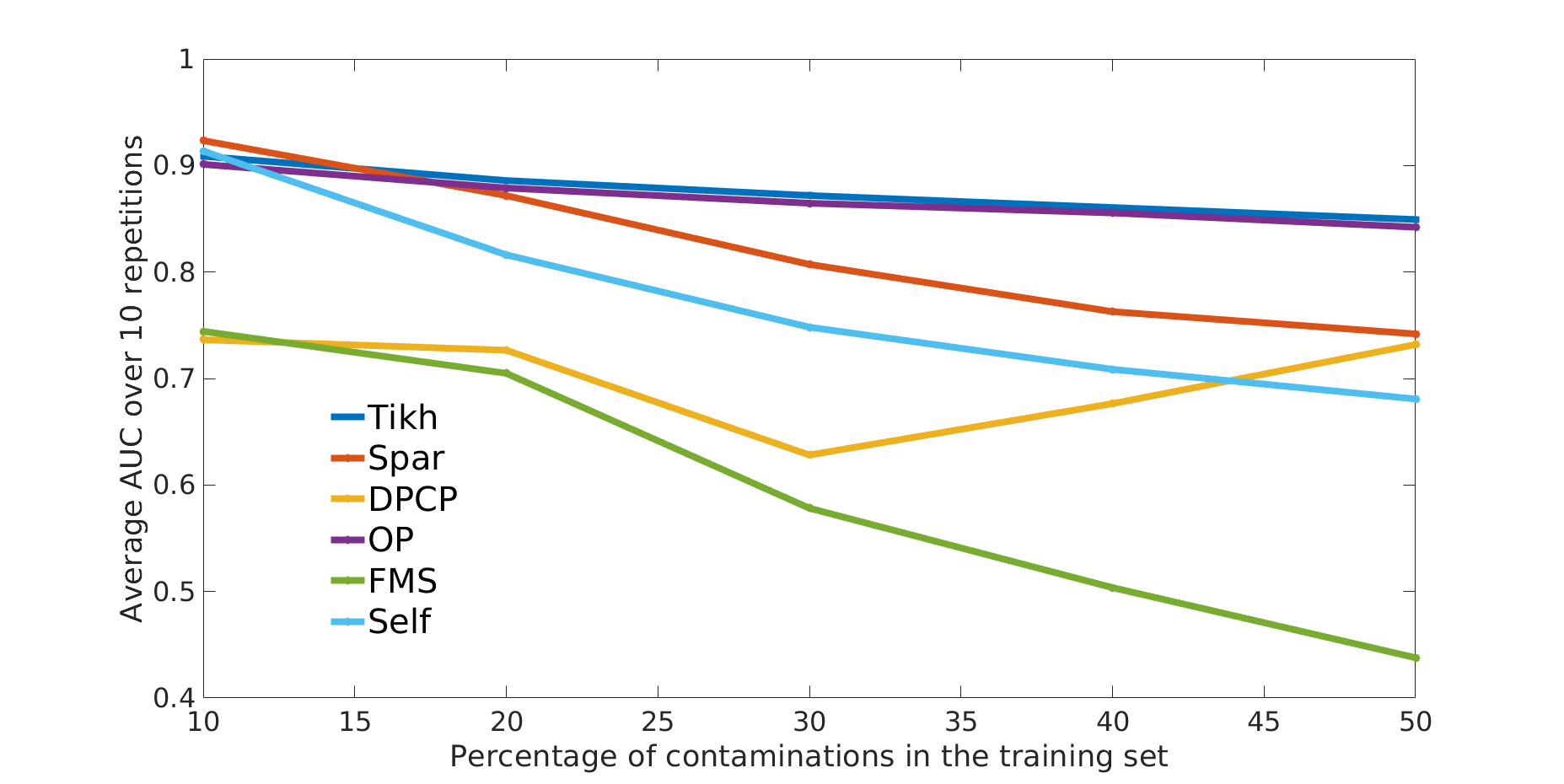}
\caption{Average AUC on the MNIST data set corresponding to a ranking of training samples.}
\label{unmnist}
\end{figure}

\begin{figure}[t]
\centering
\includegraphics[scale=.17]{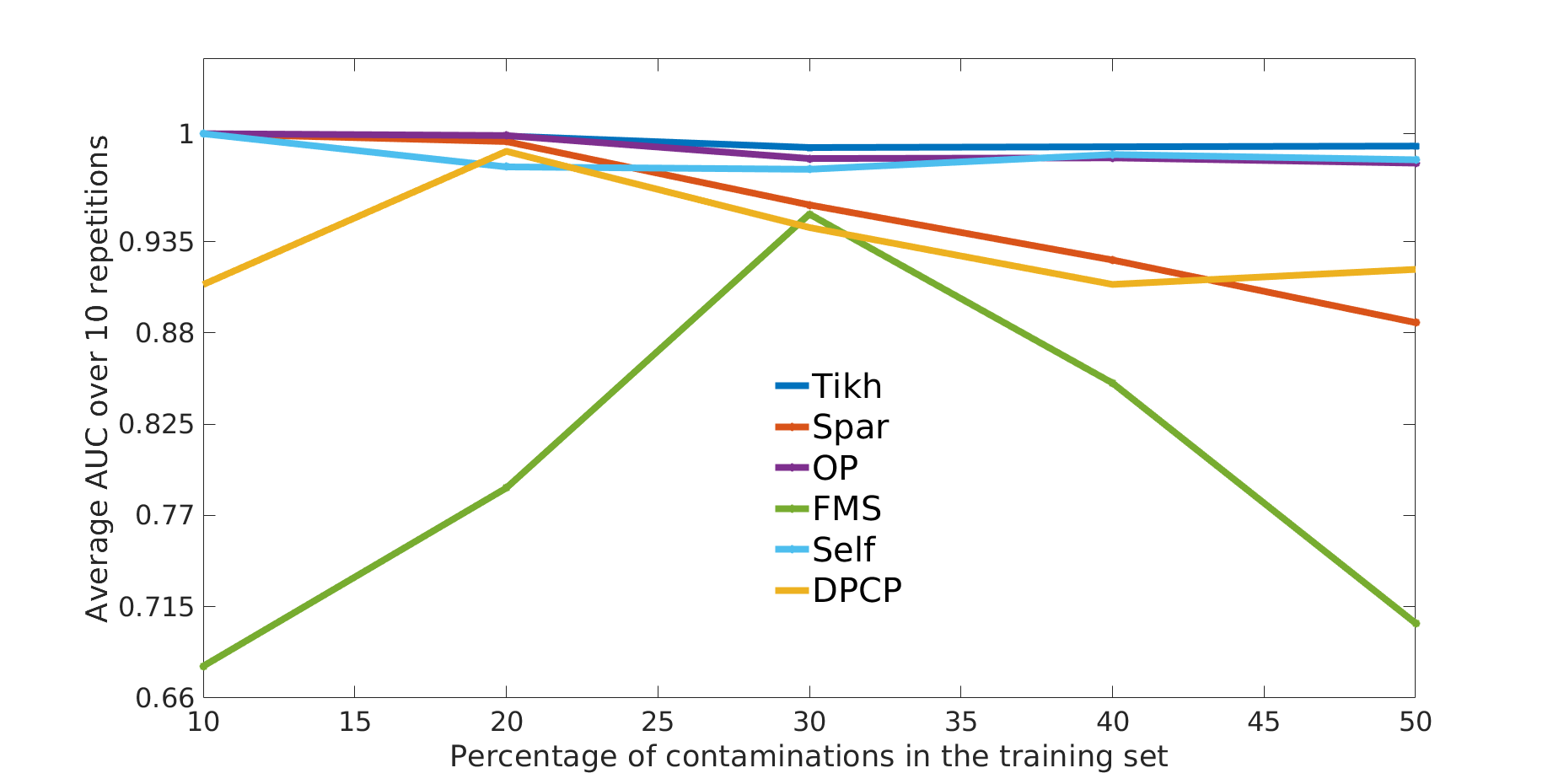}
\caption{Average AUC on the Coil-100 data set corresponding to a ranking of training samples.}
\label{uncoil}
\end{figure}

In a final set of experiments, the proposed regularised kernel spectral regression methods are evaluated in an observation ranking paradigm. As noted earlier, the original null-space kernel one-class methods are unable to provide a ranking of training observations. Nevertheless, this limitation is removed in the proposed regularised methods by imposing an additional regularisation term while at the same time updating observation label confidences via an iterative alternating minimisation technique. The methods included in the comparison in this set of experiments are the state-of-the-art methods for unsupervised observation ranking and outlier detection. Consistent with the literature \cite{Chandola:2009:ADS:1541880.1541882}, unsupervised methods refer to those approaches which do not require training data. Typically, the techniques in this category make the implicit assumption that normal instances are more frequently present than anomalies in a data set. The methods included in this experiment are those which are specifically designed to operate on a given set of contaminated samples and provide a compatibility ranking. As such, different methods in this experiment are utilised to rank observations in the training set only. The methods compared in this experiment are:
\begin{itemize}
\item \textbf{Tikh} is the proposed robust spectral regression approach using a Tikhonov regularisation term.
\item \textbf{Spar} is the proposed robust sparse spectral regression approach.
\item \textbf{DPCP} is a method for learning a linear subspace from data corrupted by outliers based on a non-convex $l_1$ optimisation problem \cite{NIPS2018_7486}. It is shown that DPCP can tolerate as many outliers as the square of the number of inliers, thus improving upon other robust PCA methods.
\item \textbf{OP} is an efficient convex optimisation-based algorithm \cite{6126034} to perform a robust principal component analysis that under mild assumptions on the uncorrupted points recovers the exact optimal low-dimensional subspace and identifies the corrupted points.
\item \textbf{FMS} is a non-convex robust subspace recovery approach \cite{doi:10.1093/imaiai/iax012}, designed to be least affected by corruptions in the training set and has been demonstrated to converge to a close vicinity of the correct subspace within few iterations
\item \textbf{SRO} obtains a weighted directed graph, defines a Markov Chain via self-representation, and identifies outliers via random walks \cite{8099943}. The SRO method can be considered as one of the leading unsupervised approaches for novelty detection.

The results, in terms of average AUC, corresponding to this experiment are provided in Fig. \ref{unface}, \ref{unmnist} and \ref{uncoil} for the face, MNIST and Coil-100 data sets, respectively and also summarised in Table \ref{unfacecoilmnistknown}. From the table and the figures, it can be observed that the Tikhonov regularisation-based approach is the top performer among other competitors on all three data sets in an unsupervised observation ranking and novelty detection scenario. The proposed sparsity-based approach performs inferior compared to the Tikhonov regression based formulation. Nevertheless, it still performs better than some other competitors including the recently proposed DPCP and FMS approaches.
\end{itemize}

\begin{table*}
\footnotesize
\renewcommand{\arraystretch}{1.2}
\caption{Summary of the average performance (in terms of AUC ($\%$)) of different methods in unsupervised observation ranking on the face, MNIST and Coil-100 data sets over the full range of contamination percentages ($10\%-50\%$).}
\label{Characteristics}
\centering
\begin{tabular}{lccccccc}
\hline
\textbf{Method} & \textbf{Tikh} & \textbf{Spar} & \textbf{DPCP} & \textbf{OP} & \textbf{FMS} & \textbf{SRO}\\
\hline
face & 96.48 & 95.34 & 73.52 & 95.98 & 72.30 & 95.85\\
MNIST & 87.52 & 82.14 & 69.99 & 86.84 & 59.38 & 77.35\\
Coil-100 & 99.50 & 95.25 & 93.39 & 99.03 & 79.44 & 98.61\\
\hline
\end{tabular}
\label{unfacecoilmnistknown}
\end{table*}

\section{Conclusion}
\label{conc}
One-class classification in a kernel Fisher null-space framework is studied. Two limitations of the null-space kernel Fisher analysis corresponding to susceptibility to a corrupted training set and inability to rank training samples are addressed. For this purpose, a regularisation of a regression-based formulation of the problem (Tikhonov and sparsity) is proposed where both projection parameters and object labels are inferred iteratively via an alternating minimisation approach. Through experiments on different data sets, it was illustrated that: 1- the proposed regularisation approach combined with the alternating optimisation mechanism is effective in robustifying the baseline method; 2-the proposed iterative ridge regression-based formulation posing one-class learning as a sensitivity analysis problem is the top performer among other competitors; and 3-the performances of both alternative regularisation schemes are boosted by automatically detecting negative samples in the training set and forming a counter-example training subset when information regarding percentage of contaminations in the training set is available. While the Tikhonov-based regularisation provides superior performance as compared with its lasso counterpart, the sparse regularisation-based method provides computational complexity advantages in the test phase since, typically, a given sample needs to be compared against only a small fraction of training samples (in the order of $10\%$ of total training data).

The proposed methodology has been evaluated in a one-class classification paradigm by assessing its generalisation capability as well as in an observation ranking scheme to detect outliers in a given data set and has been found to perform better than the baseline method while providing very competitive performances compared to the state-of-the-art techniques.


%

\ifCLASSOPTIONcompsoc
  \section*{Acknowledgments}
\else
  \section*{Acknowledgment}
\fi




%
\bibliographystyle{IEEEtran}
\bibliography{IEEEexample2}

%

\begin{IEEEbiography}[{\includegraphics[scale=.6]{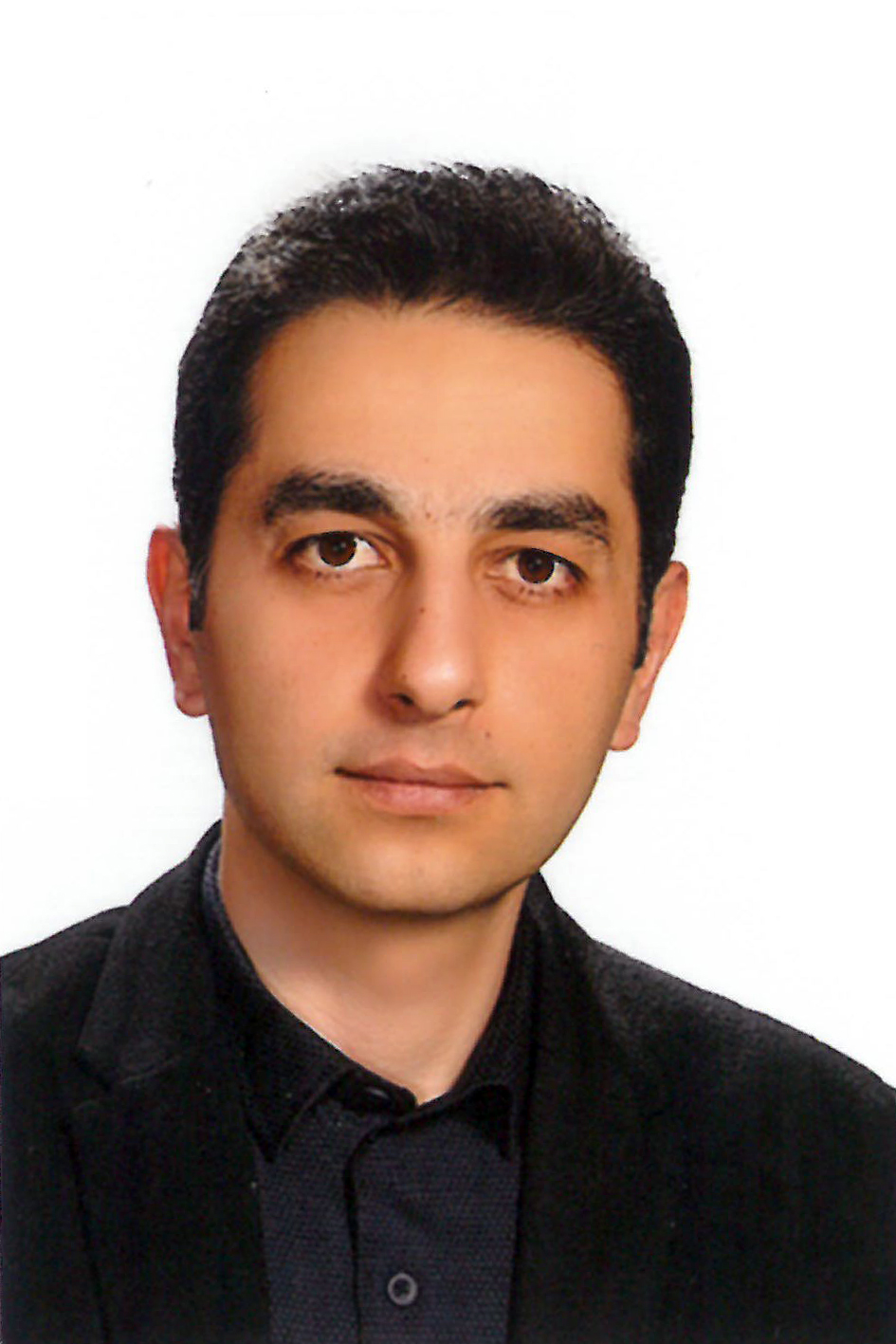}}]{Shervin Rahimzadeh Arashloo}
received the Ph.D. degree from the centre for vision, speech and signal processing, university of Surrey, UK. He is an assistant professor with the Department of Computer Engineering, Bilkent University, Ankara, Turkey and also holds a visiting research fellow position with the centre for vision, speech and signal processing, university of Surrey, UK. His research interests includes secured biometrics, novelty detection and graphical models with applications to image and video analysis.
\end{IEEEbiography}

\begin{IEEEbiography}[{\includegraphics[scale=.1]{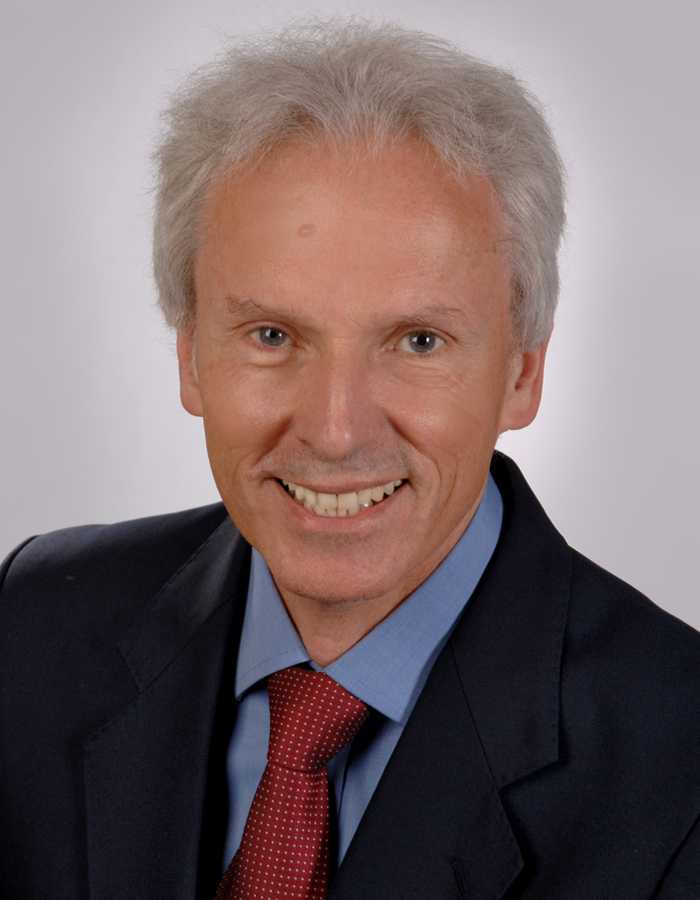}}]{Josef Kittler} (M'74-LM'12) received the B.A., Ph.D., and D.Sc. degrees from the University of Cambridge, in 1971, 1974, and 1991, respectively. He is Professor of Machine Intelligence at the Centre for Vision, Speech and Signal Processing, Department of Electronic Engineering, University of Surrey, Guildford, U.K. He conducts research in biometrics, video and image database retrieval, medical image analysis, and cognitive vision. He published the textbook Pattern Recognition: A Statistical Approach (Englewood Cliffs, NJ, USA: Prentice-Hall, 1982) and over 600 scientific papers. He serves on the Editorial Board of several scientific journals in pattern recognition and computer vision.
\end{IEEEbiography}





\end{document}